\documentclass[default]{sn-jnl}

\usepackage{graphicx}%
\usepackage{multirow}%
\usepackage{amsmath,amssymb,amsfonts}%
\usepackage{amsthm}%
\usepackage{mathrsfs}%
\usepackage[title]{appendix}%
\usepackage{xcolor}%
\usepackage{textcomp}%
\usepackage{manyfoot}%
\usepackage{booktabs}%
\usepackage{listings}%

\usepackage{booktabs}       
\usepackage{amsfonts}       
\usepackage{nicefrac}       
\usepackage{microtype}      
\usepackage{natbib}
\usepackage{doi}
\usepackage{dsfont}
\usepackage{verbatim}
\usepackage{enumerate}
\usepackage{mathtools}
\usepackage{amssymb}
\usepackage{physics} 
\usepackage{listings}
\usepackage{textcomp}
\usepackage{yhmath}
\usepackage{setspace}
\usepackage{caption}
\usepackage{float}
\usepackage{bigints}
\usepackage{pdfpages}
\usepackage{textcomp}
\usepackage{csquotes}
\usepackage{flexisym}
\usepackage{cite}
\usepackage{colortbl}
\usepackage{comment}
\usepackage{float}
\usepackage{tabularx}
\usepackage{rotating}
\usepackage{array}
\usepackage{changepage}
\usepackage{comment} 
\usepackage{longtable}
\usepackage{makecell}
\usepackage{tikz}
\usepackage{neuralnetwork}
\usepackage{xpatch}
\usepackage[ruled]{algorithm2e}
\usepackage{algpseudocode}

\setcitestyle{authoryear,open={(},close={)}}

\theoremstyle{thmstyleone}%
\theoremstyle{thmstyletwo}%

\theoremstyle{thmstylethree}%

\begin{document}

\title[Functional Autoencoder for Smoothing and Representation Learning]{Functional Autoencoder for Smoothing and Representation Learning}


\author[1]{\fnm{Sidi} \sur{Wu}}\email{sidi\_wu@sfu.ca}

\author[2]{\fnm{C\'edric} \sur{Beaulac}}\email{beaulac.cedric@uqam.ca}

\author*[1]{\fnm{Jiguo} \sur{Cao}}\email{jiguo\_cao@sfu.ca}

\affil[1]{\orgdiv{Department of Statistics and Actuarial Science}, \orgname{Simon Fraser University}, \orgaddress{\city{Burnaby}, \state{BC}, \country{Canada}}}

\affil[2]{\orgdiv{Département de Mathématiques}, \orgname{Université du Québec à Montréal}, \orgaddress{\city{Montréal}, \state{Québec}, \country{Canada}}}


\abstract{A common pipeline in functional data analysis is to first convert the discretely observed data to smooth functions, and then represent the functions by a finite-dimensional vector of coefficients summarizing the information. Existing methods for data smoothing and dimensional reduction mainly focus on learning the linear mappings from the data space to the representation space, however, learning only the linear representations may not be sufficient. In this study, we propose to learn the nonlinear representations of functional data using neural network autoencoders designed to process data in the form it is usually collected without the need of preprocessing.  We design the encoder to employ a projection layer computing the weighted inner product of the functional data and functional weights over the observed timestamp, and the decoder to apply a recovery layer that maps the finite-dimensional vector extracted from the functional data back to functional space using a set of predetermined basis functions. The developed architecture can accommodate both regularly and irregularly spaced data. Our experiments demonstrate that the proposed method outperforms functional principal component analysis in terms of prediction and classification, and maintains superior smoothing ability and better computational efficiency in comparison to the conventional autoencoders under both linear and nonlinear settings.}




\maketitle

\section{Introduction}
\label{sec:intro}
Functional data analysis (FDA) has found extensive application and received growing attention across diverse scientific domains. Functional data, as the core of FDA, is defined as any random variables that assume values in an infinite-dimensional space, such as time or spatial space in theory  \citep{fda, nonparametric_fda} , and usually discretely observed at some regularly or irregularly-spaced points over the time span in applications. Due to the complexity and difficulty in interpreting and analyzing infinite-dimensional variables, a common pipeline for functional data analysis (FDA) is to represent the infinite-dimensional functional data, denoted as $X(t)$, by a finite-dimensional vector of coefficients that extract and summarize the useful information carried by the individual functions \citep{AdaFNN}. These coefficients can be of interests themselves or be readily utilized in further analysis \citep{fda_review}. 

Two predominate approaches for dimensional reduction in FDA are basis expansion and functional principal component analysis (FPCA). The first approach, the conventional basis expansion, represents the functional data as $X_{i}(t) = \sum_{m=1}^{M_{B}}c_{im}\phi_{m}(t)$, where $\phi_{m}(t)$ are known basis functions and $c_{im}$ are corresponding basis coefficients for the $i$-th subject containing the information from the original functions \citep{fda}. This method requires the predetermination of a basis system first, for instance, Fourier or B-spline, and $M_{B}$ and the number of basis functions, to then learn the functional data's representation. The second approach, FPCA \citep{fda, nonparametric_fda, sang2017parametric}, being a fully data-driven approach, compresses the functional data $X_{i}(t)$ into functional principal component (FPC) scores $\xi_{im} = \int\{X_{i}(t)-\mu(t)\}\psi_{m}(t)dt$, where $\mu(t)$ is the mean function of the variable $X(t)$, and $\psi_{m}(t)$'s are the FPCs which are also the eigen functions derived from the spectral decomposition of the variance-covariance function of $X(t)$. By Karhunen-Lo\`eve expansion, FPCA can construct the functional data as $X_{i}(t) = \mu(t)+\sum_{j=1}^{M_{P}}\xi_{ij}\psi_{j}(t)$, with a predetermined proportion of explained variation, which indirectly defines $M_{P}$, the number of FPCs identified. The theoretical details and results on asymptotic distributions were well derived and fully discussed by \citet{dauxois1982asymptotic}, \citet{hall2006properties} and \citet{hall2006properties}. 

Representations such as FPC scores have been widely used for establishing functional regression models \citep{yao2005functional_linear_reg, Functional_additive_models, Functional_mixture_reg}, clustering \citep{Chiou2007functional_clustering, Peng2008clustering} and classification \citep{muller2005functional_classification, Muller2005generalized_flm} of functional curves. Both aforementioned methods are fundamentally linear mappings from infinite-dimensional data to the vector of finite scalars, however, learning linear projections of functional data might not be sufficient and informative. Furthermore, FPCA relies on the assumption of a common variance-covariance of all curves, which might be violated when the individual trajectories are labelled with classes.

Numerous extensions to the conventional FPCA have been suggested to adapt the linear representation of functional data for diverse scenarios \citep{yao2005functional, chen2015localized, peng2009geometric, sang2017parametric, FPCA_nonGaussian}. Nevertheless, limited contributions on nonlinear representation learning of functional data can be found in latest literature. \citet{NonlinearFPCA} extended the standard FPCA to a nonlinear additive functional principal component analysis (NAFPCA) for a vector-valued functional to accommodate nonlinear functions of functional data via two additively nested Hilbert spaces. Similar to the linear FPCA with discrete functional data, however, this technique requires to firstly estimate the underlying $X(t)$ using the basis expansion or the reproducing Kernel Hilbert space method in the first-level function space. \citet{Nonlinear_manifold_rep} developed nonlinear manifold learning to generate nonlinear representations of functional data by modifying the existing nonlinear dimension reduction methods to satisfy functional data settings. The manifold-based representation is basically designed to be layered on the representation produced by FPCA, while its computational difficulties may arise as the sample size increases. 

Meanwhile, the advent use of big data and gradual popularity of deep learning promote the introduction of neural networks to functional data representation learning. \citet{FDA_nonlinar_learning} explored a functional nonlinear learning (FunNoL) method relying on recurrent neural networks (RNN) to represent multivariate functional data in a lower-dimensional feature space and handle the missing observations and excessive local disturbances of observed functional data. This method ignores the basic structure of functional data as it regards $X(t)$ as time series data and captures the temporal dependency across time sequences. Moreover, to enable the usage of representation in classifying curves, FunNoL is designed to be a semi-supervised model that merges a classification model to a standard RNN, introducing more complexity in network optimization and representation learning. \citet{Functional_AE} defined a functional autoencoders, which generalize the conventional neural network autoencoders to handle continuous functional data, and derived the functional gradient-based learning algorithm for autoencoder optimization, to study the nonlinear projection of multidimensional functional data. This approach requires smooth functional inputs and overlooks the common issue where functional data are barely fully observed in practice \citep{yao2005functional}. 

The main objective of this study is to propose a solution to the nonlinear representation learning and smoothing of discrete functional data using a novel functional autoencoders (FAE) based on dense feed-forward neural networks, which include FPCA as a special case under the linear representation setting. As an unsupervised learning technique, autoencoders (AE) have been frequently used for feature extraction and representation learning in vector-space problem \citep{hinton2006reducing, wang2016auto, Meiler2001-be, BengioY2013RLAR}. A traditional AE consists of an encoder and a decoder connected by a bottleneck layer, where the former one is a mapping from a $P$-dimensional vector-valued input space to a $d$-dimensional representation space and the latter one maps from the $d$-dimensional representation space back to a vector-valued output space of $P$ dimensions, where the output layer consists of a reconstruction of the original input. Assuming  $d << P$, the neurons in the bottleneck layer serve as a lower dimension representation of the input. This representation is a collection of neuron features extracted from the AE and can be of interests themselves or can be used for further research. The relation between AE and principal component analysis (PCA) has been well discussed in several existing studies. \citet{Oja1, Oja2} demonstrated that a neural network employing a linear activation function essentially learns the principal component representation of the input data. Furthermore, \citet{BaldiPierre1989Nnap} and \citet{BengioY2013RLAR} demonstrated that an autoencoder with one hidden layer and identity activation is essentially equivalent to PCA. \citet{Bourlard1988-fk} and \citet{BaldiPierre1989Nnap} also explained that the representations captured by autoencoders are a basis of the subspace spanned by the leading principal components (PCs) instead of necessarily coincident with them. The connection between the conventional AE and the PCA can be naturally transplanted to that of the designed FAE and the FPCA with a relevant discussion provided.

In this work, we propose to construct an autoencoder under discrete functional data settings; a functional autoencoder. 
We design the encoder to incorporate a projection layer computing the weighted inner product of the functional data and functional weights over the observed discrete time span, and the decoder to equip a recovery layer to map the finite-dimensional vector extracted from the functional data to functional space using a set of pre-selected basis functions. The developed architecture compresses the discretely observed functional data to a set of representations and then outputs smooth functions. The resulting lower-dimensional vector will be the representation/encoding of the functional data, serving a similar purpose to the basis coefficients or FPC scores previously mentioned and can be inputted into any further analysis. 

The autoencoder we designed for functional data have at least the following highlights. First, the proposed FAE addresses the learning of a nonlinear representation from discrete functional data with a flexible nonlinear mapping path captured by neural networks, which eliminates the conduct of curve smoothing assuming any particular form in advance. As a result, our method performs a one-step model simultaneously learning the representative feature and smoothing the discretely observed trajectories. Second, it allows us to obtain linear and nonlinear projections of functional data, with the former path serving as an alternative approach to FPCA. Third, the proposed method is applicable for both regularly and irregularly spaced data, while the smoothness of the recovered curves is controlled through a roughness penalty added to the objective function in training. Forth, the architecture of the FAE is flexibly programmable and compatible with existing neural networks libraries/modules. Last but not the least, the robustness and efficiency of our method in representation extraction and curve recovery with small size of data and substantial missing information are supported by the results of various numerical experiments. 

The remainder of this article proceeds in the following manner. In Section \ref{sec:FAE}, we provide the methodological details for the proposed FAE, including a description about the network architecture and an explanation on the corresponding training procedure. A brief discussion on the connections between the proposed method and two well-established methods, FPCA and AE, is given in Section \ref{sec:connection}. In Section \ref{sec:simulation}, we compare the proposed functional autoencoders to the existing methods applicable for functional data representation with a focus on relationship capture and computational efficiency via intensive simulation studies under various scenarios. The designed autoencoders and other techniques in comparison are further evaluated in Section \ref{sec:real_application} under a real data application. Finally, we conclude with a discussion and future directions in Section \ref{sec:conclusion}. The pre-processed data sets and computing codes of the proposed method on selected applications are available at \url{https://github.com/CedricBeaulac/FAE}.

\section{Functional autoencoders}
\label{sec:FAE}

\subsection{Motivation: autoencoders for continuous functional data}
\label{sec_CFAE}
Suppose there are $N$ subjects and for the $i$-th subject, a functional variable $X_{i}(t), t \in \mathcal{T}$ is observed in the $L^{2}(t)$ space. To address the limitations of linear representations of functional data $X(t)$, we propose to learn nonlinear mappings from functional data space $L^{2}(t)$ to $K$-dimensional vector space $\mathbb{R}^{K}$ through a neural network autoencoder which contains an encoder compressing the functional input to some scalar-valued neurons, and a decoder reconstructing the functional input back from the encoded representations. 

We introduce an autoencoder with $L$ hidden layers (excluding the input and output layers) for continuous functional data $X(t)$, which we suppose, are  fully observed over a continuum $t$. Different from conventional autoencoders consuming scalar inputs, in this scenario, functions are served as inputs and fed into the neural network, and the designed autoencoder for continuous functional data is supposed to be trained by minimizing the reconstruction error $L(X(t), \hat{X}(t)) = \frac{1}{N_{\text{train}}} \sum_{i=1}^{N_{\text{train}}}\int_{\mathcal{T}} (X_i(t)-\hat{X}_i(t))^2 dt$.

We propose to encode the infinite-dimensional functions to some finite number of numerical neurons by introducing functional weights $w^{I}(t)$ for bridging the input and the first hidden layer of the encoder. Specifically, the scalar inner product, which connects neurons in the input and the first layers of the conventional AE, is generalized by the inner product of the functional input $X(t)$ and functional weight $w^{I}(t)$ in $L^{2}$ space. Consequently, the $k$-th neuron in the first hidden layer $h_{k}^{(1)}$ is computed as 
\begin{align}
     h_{k}^{(1)}&=g\left(\int_{\mathcal{T}} X(t)w_{k}^{I}(t)dt\right),
     \label{eq_CFAE_input}
\end{align}
where $w_{k}^{I}(t)$ is the input functional weight connecting the functional input and the $k$-th neuron in the first hidden layer, and $g(\cdot)$ is the activation function. To be noted that here we opted to neglect the numerical bias term $b^{(1)}$ for simplicity. 

The proposed functional weights together with the inner product of two functions achieve the mapping from $L
^{2}$ to $\mathbb{R}^{K^{(1)}}$, where $K^{(l)}$ is the number of neurons in the $l$-th hidden layer and $l \in \{1, 2,...,L\}$. The resulting numerical neurons are further passed to the continuous hidden layers of the autoencoders, following the same calculation rules as in conventional AEs. Accordingly, the $k$-th neuron in the $l$-th hidden layers is given by
\begin{align}
    h_{k}^{(l)} = g\left(\sum_{j}^{K^{(l)}}h_{j}^{(l-1)}w_{j}^{(l)}\right),
\label{eq_CFAE_input_hidden}
\end{align}
with $h^{l-1}_{j}$ being the $j$-th neuron in the $l-1$ layer connected by the scalar network weight $w_{j}^{(l)}$. 

Similarly, a set of functional weights $\{w_{k}^{O}(t)\}_{k=1}^{K^{(L)}}$, instead of scalar weights, are applied at the output layer of the decoder to map the second to last layer from $\mathbb{R}^{K^{(L)}}$ back to functional space $L^{2}$ and mathematically, the outputted functional neuron is calculated as
\begin{align}
    \hat{X}(t)&=\sum_{k=1}^{K^{(L)}} h_{k}^{(L)}w_{k}^{(O)}(t),
\label{eq_CFAE_output}
\end{align}
where $w_{k}^{O}(t)$ is the output functional weight connecting the $k$-th neuron in the $L$-th hidden layer and the functional neuron. For this output layer to produce the functional output required, the linear activation function must be used. 

\begin{figure}
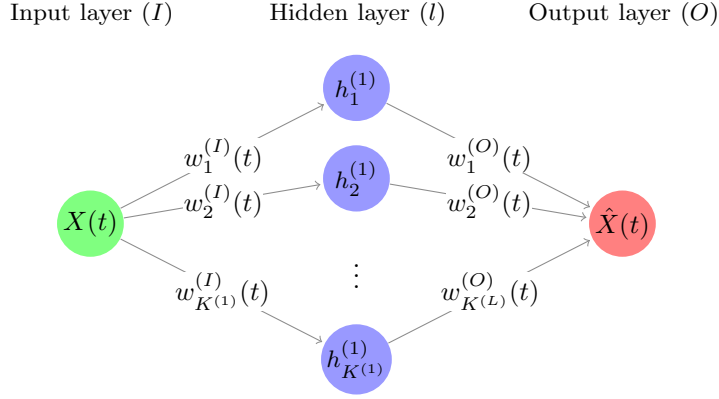

\begin{center}
\begin{neuralnetwork}[nodespacing=12mm, layerspacing=35mm,
   maintitleheight=2.5em, layertitleheight=4.5em,
   height=3, toprow=false, nodesize=25pt, style={},title={}, titlestyle={}]
    \newcommand{\x}[2]{$X(t)$}
    \newcommand{\xhat}[2]{$\hat{X}(t)$}
    \newcommand{\hfirst}[2]{\scriptsize \ifnum #2=4 $h^{(1)}_{K^{(1)}}$ \else $h^{(1)}_#2$ \fi}
    \newcommand{\hsecond}[2]{\small $h^{(2)}_#2$}
    \newcommand{\wfirst}[4]{\ifnum #4=4 $w^{(I)}_{K^{(1)}}(t)$ \else $w^{(I)}_{#4}(t)$ \fi}
    \setdefaultlinklabel{\wfirst}
    \inputlayer[count=1, bias=false, text=\x, title={\small Input layer ($I$)}]
    \hiddenlayer[count=4,  exclude={3}, bias=false, text=\hfirst, title= {\small Hidden layer ($l$)}] 
        \foreach \n in {1}{
       \foreach \m in {1,2,4}{
        \link[style={}, labelpos=midway, from layer=0, from node=\n, to layer=1, to node=\m]
        }
    }
    \newcommand{\wsecond}[4]{\ifnum #2=4 $w^{(O)}_{K^{(L)}}(t)$ \else $w^{(O)}_{#2}(t)$ \fi}
    \setdefaultlinklabel{\wsecond}
    \outputlayer[count=1, text=\xhat, title={\small Output layer ($O$)}] 
        \foreach \n in {1,2,4}{
       \foreach \m in {1}{
        \link[style={}, labelpos=midway, from layer=1, from node=\n, to layer=2, to node=\m]
        }
    }
    \path (L1-2) -- node{$\vdots$} (L1-4);
\end{neuralnetwork}
\vspace{1em}
\caption{Functional autoencoder for continuous data with $L=1$ hidden layer.}
\label{fig_CFAE}
\end{center}
\end{figure}

We name this autoencoder the continuous functional autoencoder (CFAE) and a graphical visualization of the CFAE with $L=1$ hidden layer(s) for functional data can be seen in Figure \ref{fig_CFAE}. In this scenario, $\{h_{1}^{(1)},h_{2}^{(1)},..., h_{K^{(1)}}^{(1)}\}$ is regarded as the vector-valued representation of $X(t)$.

\subsection{Proposed model: autoencoders for discrete functional data}
\label{sec_DFAE}


The CFAE introduced in Section \ref{sec_CFAE} serves as an inspiration for the model we proposed better suited for discrete functional data, which is how functional data are collected and stored in practical applications. Considering the functional data are discretely observed at $J$ evenly spaced time points $(t_1,...,t_J)$ over the time interval $\mathcal{T}$ for all $N$ subjects, and therefore for the $i$-th subject, we obtain $J$ pairs of observations $\{t_{j}, X_{i}(t_{j})\}$, $j = 1, 2, ... J$. As a matter of fact, the real functional data are often contaminated with some observational errors, resulting in a collection of noisy discrete observations $\tilde{X}_{i}(t_{j}) = X_{i}(t_{j}) + \epsilon_{i}(t_{j})$, where $\epsilon_{i}(t_{j})$ is the i.i.d. measurement error. Without knowing the true underlying curves $X(t)$'s, the contaminated observations $\{\tilde{X}(t_{j})\}_{j=1}^{J}$'s are employed as an alternative to $\{X(t_{j})\}_{j=1}^{J}$'s in applications.

We propose to adapt the CFAE to take data $\{X(t_1), X(t_2), ..., X(t_J)\}$, a $J$-dimensional vector, as the input. Instead of smoothing the discrete data and then applying the previously defined autoencoder, we develop the architecture to satisfy such discrete functional input. This is a major advantage of our proposed method as the data no longer needs to be preprocessed in any way before being fed to our proposed FAE. To do so, we replace the weight functions $w_{k}^{I}(t)$ for the input layer and $w_{k}^{O}(t)$ for the output layer with their discrete versions $\{w_{k}^{I}(t_j)\}_{j=1}^{J}$ and $\{w_{k}^{O}(t_j)\}_{j=1}^{J}$ evaluated at the corresponding $J$ time points $(t_1,...,t_J)$, respectively. Naturally, the integral $\int_{\mathcal{T}} X(t)w_{k}^{(I)}(t)dt$ in Eq.(\ref{eq_CFAE_input}) is approximated numerically using the rectangular or trapezoidal rule and accordingly the $k$-th neuron in the first hidden layer is updated as
\begin{align}
    h_{k}^{(1)}&=g\left(\sum_{j=1}^{J} \omega_{j} X(t_{j})w_{k}^{(I)}(t_{j})\right),
\label{eq_DFAE_input}
\end{align}
where $\{\omega_{j}\}_{j=1}^{J}$ are the weights used in the numerical integration algorithm. 

Likewise, the output layer now consists of $J$ neurons corresponding to the $J$-dimensional input vector as
\begin{align}
    \hat{X}(t_{j})&=\sum_{k=1}^{K^{(L)}} h_{k}^{(L)}w_{k}^{(O)}(t_{j}).
\label{eq_DFAE_output}
\end{align}

\begin{figure}
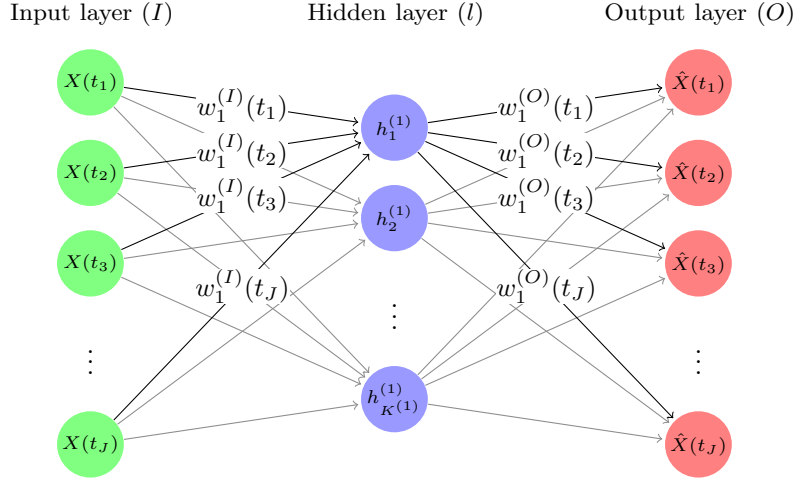

\centering
\begin{neuralnetwork}[nodespacing=12mm, layerspacing=40mm,
   maintitleheight=1.5em, layertitleheight=6em,
   height=3, toprow=false, nodesize=25pt, style={},title={}, titlestyle={}]
    \newcommand{\x}[2]{\footnotesize \ifnum #2=5 $X(t_{J})$ \else $X(t_#2)$ \fi}
    \newcommand{\xhat}[2]{\footnotesize \ifnum #2=5 $\hat{X}(t_{J})$ \else $\hat{X}(t_#2)$ \fi}
    \newcommand{\hfirst}[2]{\footnotesize \ifnum #2=4 $h^{(1)}_{K^{(1)}}$ \else $h^{(1)}_#2$ \fi}
    \newcommand{\hsecond}[2]{\footnotesize $h^{(2)}_#2$}

    \inputlayer[count=5, exclude={4}, bias=false, text=\x, title={\small Input layer ($I$)}]
    \hiddenlayer[count=4, exclude={3}, bias=false, text=\hfirst,title={\small Hidden layer ($l$)}] 
    \foreach \n in {1, 2, 3, 5}{
       \foreach \m in {2, 4}{
        \link[style={}, labelpos=midway, from layer=0, from node=\n, to layer=1, to node=\m]
        }
    }
    \newcommand{\wfirst}[4]{\ifnum #2=5 $w^{(I)}_{#4}(t_{J})$ \else $w^{(I)}_{#4}(t_#2)$ \fi}
    \setdefaultlinklabel{\wfirst}
    \foreach \n in {1, 2, 3, 5}{
        \foreach \m in {1}{
            \link[style={}, labelpos=midway, from layer=0, from node=\n, to layer=1, to node=\m, style=black]
        }
    }

    \outputlayer[count=5, exclude={4}, text=\xhat, title={\small Output layer ($O$)}] 
    \newcommand{\wnone}[4]{}
    \setdefaultlinklabel{\wnone}
    \foreach \n in {2,4}{
       \foreach \m in {1, 2, 3, 5}{
        \link[style={}, labelpos=midway, from layer=1, from node=\n, to layer=2, to node=\m]
        }
    }
    \newcommand{\wsecond}[4]{\ifnum #4=5 $w^{(O)}_{#2}(t_J)$ \else $w^{(O)}_{#2}(t_#4)$ \fi}
    \setdefaultlinklabel{\wsecond}
    \foreach \n in {1}{
        \foreach \m in {1, 2, 3, 5}{
            \link[style={}, labelpos=midway, from layer=1, from node=\n, to layer=2, to node=\m, style=black]
        }
    }
    \path (L0-3) -- node{$\vdots$} (L0-5);
    \path (L1-2) -- node{$\vdots$} (L1-4);
    \path (L2-3) -- node{$\vdots$} (L2-5);
\end{neuralnetwork}
\vspace{1em}
\caption{Functional autoencoder for discrete data with $L=1$ hidden layer.}
\label{fig_AE_discrete}
\end{figure}

As illustrated in Figure \ref{fig_AE_discrete}, the mappings $L^{2} \rightarrow \mathbb{R}^{K^{(1)}}$ and $\mathbb{R}^{K^{(L)}} \rightarrow L^{2}$ in the CFAE are substituted with $\mathbb{R}^{J} \rightarrow \mathbb{R}^{K^{(1)}}$ and $\mathbb{R}^{K^{(L)}} \rightarrow \mathbb{R}^{J}$, respectively, for this discrete setting. 

The autoencoder for discrete functional data seemingly behaves the same as a conventional autoencoder, however, the proposed FAE requires a different training process which accounts for the assumption that functional data are the realization of a underlying smooth stochastic process. This way, the proposed FAE considers the serial correlation of the functional data and returns a smooth and continuous functional data without the need of smoothing in the input preemptively.

We propose to represent functional weight as $w^{(\cdot)}_k(t) = \sum_{m=1}^{M_k^{(\cdot)}}c^{(\cdot)}_{mk}\phi^{(\cdot)}_{mk}(t)$, where $\{\phi_{k}^{(\cdot)}(t)\}_{m=1}^{M_{k}^{(\cdot)}}$'s are some known basis functions from a selected basis system, such as Fourier or B-spline, $\{c_{mk}^{(\cdot)}\}_{m=1}^{M_{k}^{(\cdot)}}$ are the corresponding basis coefficients remain to be determined, and $M_{k}^{(\cdot)}$ is some predefined truncation integer for the $k$-th weight. The snapshots of the $k$-th input or output weight function are accordingly marked as
\begin{align}
    w_{k}^{(\cdot)}(t_j) = \sum_{m=1}^{M_k^{(\cdot)}}c^{(\cdot)}_{mk}\phi^{(\cdot)}_{mk}(t_j),
\end{align}
and therefore Eq.(\ref{eq_DFAE_input}) and Eq.(\ref{eq_DFAE_output}) calculating the $k$-th neurons in the first hidden layer and the output layer, respectively, can be re-written as
\begin{align}
    h_{k}^{(1)} &=g\left(\sum_{j=1}^{J} \omega_{j}X(t_{j})\sum_{m=1}^{M_k^{(I)}}c^{(I)}_{mk}\phi^{(I)}_{mk}(t_j) \right) \notag \\   &=g\left(\sum_{m=1}^{M_k^{(I)}}c^{(I)}_{mk}\sum_{j=1}^{J}\omega_{j}X(t_{j})\phi^{(I)}_{mk}(t_j)\right), \label{eq_DFAE_input_approx}\\
     \hat{X}(t_{j})&=\sum_{k=1}^{K^{(L)}} h_{k}^{(L)}\sum_{m=1}^{M_k^{(O)}}c^{(O)}_{mk}\phi^{(O)}_{mk}(t_{j}) \notag  \\
    &=\sum_{k=1}^{K^{(L)}}\sum_{m=1}^{M_k^{(O)}} h_{k}^{(L)}c^{(O)}_{mk}\phi^{(O)}_{mk}(t_{j}).
    \label{eq_DFAE_output_approx}
\end{align}

In such a manner, the problem of learning a functional weight $w^{(\cdot)}_k(t)$ using typical machine learning techniques  becomes one of learning $\{c_{mk}^{(\cdot)}\}_{m=1}^{M_{k}^{(\cdot)}}$, the parameters defining $w^{(\cdot)}_k(t_j)$, for $k=1,...,K^{(l)}$, $l=1 \text{ or } O$, and consequently, we seek to learn the coefficients $\{c_{mk}^{(\cdot)}\}_{m=1}^{M_{k}^{(\cdot)}}$ through back-propagation. 

\subsubsection{Encoder with a feature layer}
\label{sec_encoder_w/_feature_layer}

\begin{figure}
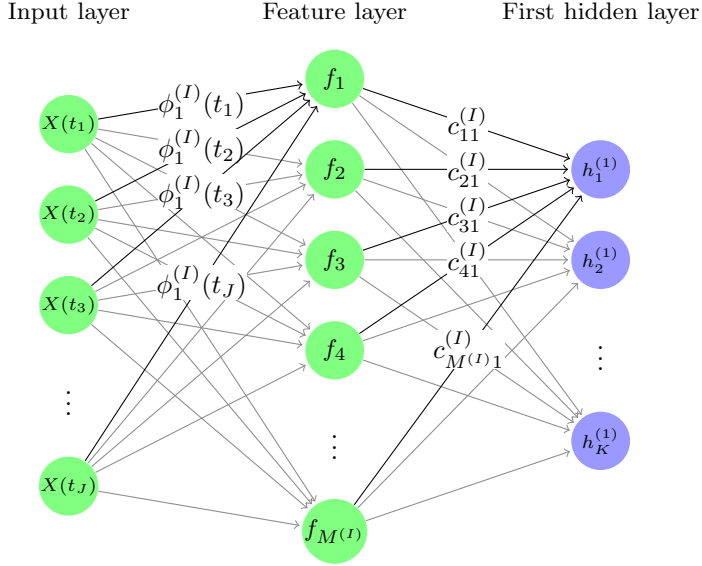

\centering
\begin{neuralnetwork}[nodespacing=12mm, layerspacing=35mm,
   maintitleheight=1em, layertitleheight=11em,
   height=1, toprow=false, nodesize=22pt, style={},title={}, titlestyle={}]
    \newcommand{\x}[2]{\footnotesize \ifnum #2=5 $X(t_{J})$ \else $X(t_#2)$ \fi}
    \newcommand{\f}[2]{\small \ifnum #2=6 $f_{M^{(I)}}$ \else $f_#2$ \fi}
    \newcommand{\fhat}[2]{\small \ifnum #2=6 $b_{^{(O)}M}$ \else $b_#2$ \fi}
    \newcommand{\xhat}[2]{\footnotesize \ifnum #2=5 $\hat{X}(t_{J})$ \else $\hat{X}(t_#2)$ \fi}
    \newcommand{\hfirst}[2]{\footnotesize \ifnum #2=4 $h^{(1)}_{K}$ \else $h^{(1)}_#2$ \fi}
    \newcommand{\hsecond}[2]{\footnotesize $h^{(2)}_#2$}

    \inputlayer[count=5, exclude={4}, bias=false, text=\x, title = {\small Input layer}]
    \inputlayer[count=6, exclude={5}, bias=false, text=\f, title = {\small Feature layer}] 
    \foreach \n in {1,2,3,5}{
       \foreach \m in {2, 3, 4, 6}{
        \link[style={}, labelpos=midway, from layer=0, from node=\n, to layer=1, to node=\m]
        }
    }
    \newcommand{\wfirst}[4]{\ifnum #2=5 $\phi_{#4}^{(I)}(t_{J})$ \else $\phi_{#4}^{(I)}(t_#2)$ \fi}
    \setdefaultlinklabel{\wfirst}
    \foreach \n in {1,2,3,5}{
        \foreach \m in {1}{
            \link[style={}, labelpos=midway, from layer=0, from node=\n, to layer=1, to node=\m, style=black]
        }
    }

    \hiddenlayer[count=4, bias=false, exclude={3}, text=\hfirst, title={\small First hidden layer}]
    \newcommand{\wsecond}[4]{\ifnum #2=6 $c^{(I)}_{M^{(I)}1}$ \else $c^{(I)}_{#21}$ \fi}
     \newcommand{\wnone}[4]{}
    \setdefaultlinklabel{\wnone}
    \foreach \n in {1,...,4,6}{
       \foreach \m in {2,4}{
        \link[style={}, labelpos=midway, from layer=1, from node=\n, to layer=2, to node=\m]
        }
    }
    \setdefaultlinklabel{\wsecond}
    \foreach \n in {1,...,4,6}{
       \foreach \m in {1}{
        \link[style={}, labelpos=midway, from layer=1, from node=\n, to layer=2, to node=\m, style=black]
        }
    }
    \path (L0-3) -- node{$\vdots$} (L0-5);
    \path (L1-4) -- node{$\vdots$} (L1-6);
    \path (L2-2) -- node{$\vdots$} (L2-4);
\end{neuralnetwork}
\vspace{1em}
\caption{Encoder with a Feature Layer. Notice that the input and feature layers are devoid of parameters at this point and are entirely deterministic given the data and the choice of basis function for $w$.}
\label{fig_encoder_w/_feature_layer}
\end{figure}

For computational convenience, we let $M_k^{(I)} = M^{(I)}$ and $\phi_{mk}^{(I)}(t) = \phi_{m}^{(I)}(t)$ for all $k \in \{1, 2, ..., K^{(1)}\}$, indicating that the input weight functions $\{w_{k}^{(I)}(t)\}_{k=1}^{K^{(1)}}$ are expressed with the same basis expansion. In consequence, we can simplify Eq. (\ref{eq_DFAE_input_approx}) as
 \begin{align}
       h_{k}^{(1)} 
       &=g\left(\sum_{m=1}^{M^{(I)}}c^{(I)}_{mk}\sum_{j=1}^{J}\omega_{j}X(t_{j})\phi^{(I)}_{m}(t_j)\right) \notag \\
       &=g\left(\sum_{m=1}^{M^{(I)}}c^{(I)}_{mk}f_{m}\right),
\end{align}
where $f_{m} =\sum_{j=1}^{J}\omega_{j}X(t_{j})\phi^{(I)}_{m}(t_{j}), m = \{1, 2, ..., M^{(I)}\}$, a Riemann sum approximating the inner product of $X(t)$ and $\phi^{(I)}_{m}(t)$. $\{f_{m}\}_{m=1}^{M^{(I)}}$ represent the resulting \textit{features} of $X(t)$ projected to the basis function sets and serve as the pivot connecting the input layer and the first hidden layer. Hence, we design our proposed encoder by inserting a \textit{feature layer} of $f_{m}$'s between the input layer and the first hidden layer, as shown in Figure \ref{fig_encoder_w/_feature_layer}. This deterministic layer translates discretely observed functional data into a scalar structure that can then be processed with existing neural network models and training algorithms.

Specifically, the input layer and the \textit{feature layer} are linked by the snapshots of the basis function set $\{\phi^{(I)}_{m}(t_{j})\}_{m=1}^{M^{(I)}}$, while $c^{(I)}_{mk}$ becomes the network weights connecting the \textit{feature layer} and the first hidden layer.

A benefit of this layer architecture is that different observations that might be observed at different time points will all get converted to the same features. Consequently, irregularly observed data are managed through this deterministic layer. It is also possible to recover the continuous functional weights $\{w_{k}^{(I)}(t)\}_{k=1}^{K^{(1)}}$ for visualization purpose after learning the coefficients $\{c_{mk}^{(\cdot)}\}_{m=1}^{M_{k}^{(\cdot)}}$ through training. 

\subsubsection{Decoder with a coefficient layer}
\label{sec_decoder_w/_coef_layer}

Again, for simplicity, we set the basis functions to be the same for representing all output weight functions $\{w_{k}^{(O)}(t)\}_{k=1}^{K^{(L)}}$ by setting $M_k^{(O)} = M^{(O)}$ and $\phi_{mk}^{(O)}(t) = \phi_{m}^{(O)}(t)$ for all $k \in \{1, 2, ..., K^{(L)}\}$. Following on Eq.(\ref{eq_DFAE_output_approx}), we now have:
\begin{align}
    \hat{X}(t_{j}) 
    & =\sum_{k=1}^{K^{(L)}}\sum_{m=1}^{M^{(O)}} h_{k}^{(L)}c^{(O)}_{mk}\phi^{(O)}_{m}(t_{j}) \notag \\
    & =\sum_{m=1}^{M^{(O)}}\left(\sum_{k=1}^{K^{(L)}} h_{k}^{(L)}c^{(O)}_{mk}\right)\phi^{(O)}_{m}(t_{j}) \notag \\
    & = \sum_{m=1}^{M^{(O)}}b_{m}\phi^{(O)}_{m}(t_{j}),
\end{align}
where $b_{m} = \sum_{k=1}^{K^{(L)}}h_{k}^{(L)}c^{(O)}_{mk}$. In fact, $\{\phi_{m}^{(O)}(t)\}_{m=1}^{M^{(O)}}$ can be regarded as the basis functions used in the representation of $\hat{X}(t)$, the reconstructed functional observation. In turn, $\{b_{m}\}_{m=1}^{M^{(O)}}$ play the role of the corresponding basis coefficients. 

\begin{figure}
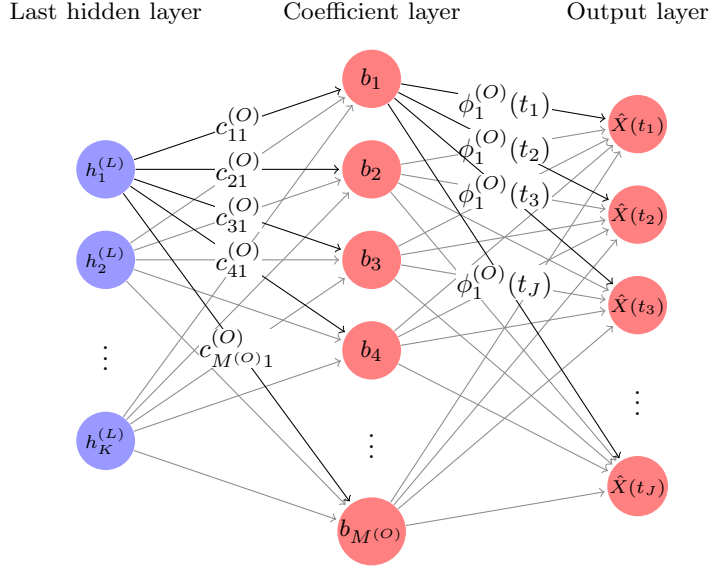

\centering
\begin{neuralnetwork}[nodespacing=12mm, layerspacing=35mm,
   maintitleheight=1em, layertitleheight=11em,
   height=1, toprow=false, nodesize=22pt, style={},title={}, titlestyle={}]
    \newcommand{\x}[2]{\footnotesize \ifnum #2=5 $x(t_{J})$ \else $x(t_#2)$ \fi}
    \newcommand{\f}[2]{\small \ifnum #2=6 $f_{M^{(I)}}$ \else $f_#2$ \fi}
    \newcommand{\fhat}[2]{\small \ifnum #2=6 $b_{M^{(O)}}$ \else $b_#2$ \fi}
    \newcommand{\xhat}[2]{\footnotesize \ifnum #2=5 $\hat{X}(t_{J})$ \else $\hat{X}(t_#2)$ \fi}
    \newcommand{\hfirst}[2]{\footnotesize \ifnum #2=4 $h^{(L)}_{K}$ \else $h^{(L)}_#2$ \fi}
    \newcommand{\hsecond}[2]{\footnotesize $h^{(2)}_#2$}


    \hiddenlayer[count=4, bias=false, exclude={3}, text=\hfirst, title={\small Last hidden layer}]
    \outputlayer[count=6, bias=false,exclude={5}, text=\fhat, title={\small Coefficient layer}]
    \foreach \n in {2,4}{
       \foreach \m in {1,...,4,6}{
        \link[style={}, labelpos=midway, from layer=0, from node=\n, to layer=1, to node=\m]
        }
    }
    \newcommand{\wfirst}[4]{\ifnum #4=6 $c^{(O)}_{M^{(O)}1}$ \else $c^{(O)}_{#41}$ \fi}
    \setdefaultlinklabel{\wfirst}
    \foreach \n in {1}{
       \foreach \m in {1,...,4,6}{
        \link[style={}, labelpos=midway, from layer=0, from node=\n, to layer=1, to node=\m, style=black]
        }
    }
  
    \outputlayer[count=5, exclude={4}, text=\xhat, title ={\small Output layer}] 
    \newcommand{\wnone}[4]{}
    \setdefaultlinklabel{\wnone}
    \foreach \n in {2,...,4,6}{
       \foreach \m in {1,...,3,5}{
        \link[style={}, labelpos=midway, from layer=1, from node=\n, to layer=2, to node=\m]
        }
    }
    \newcommand{\wsecond}[4]{\ifnum #4=5 $\phi_{#2}^{(O)}(t_{J})$ \else $\phi_{#2}^{(O)}(t_#4)$ \fi}
    \setdefaultlinklabel{\wsecond}
    \foreach \n in {1}{
        \foreach \m in {1,...,3,5}{
            \link[style={}, labelpos=midway, from layer=1, from node=\n, to layer=2, to node=\m, style=black]
        }
    }
    \path (L0-2) -- node{$\vdots$} (L0-4);
    \path (L1-4) -- node{$\vdots$} (L1-6);
    \path (L2-3) -- node{$\vdots$} (L2-5);
\end{neuralnetwork}
\vspace{1em}
\caption{Decoder with a Coefficient Layer. Similarly, the last two layers are devoid of parameters and are deterministic.}
\label{fig_decoder_w/_coef_layer}
\end{figure}

Hence, in Figure \ref{fig_decoder_w/_coef_layer}, we visualize $b_{m}$'s as the neurons of a \textit{coefficient layer} added to the decoder for connecting the last hidden layer and the output layer, while $c^{(O)}_{mk}$ are the network weights between the last hidden layer and the \textit{coefficient layer}, and meanwhile the \textit{coefficient layer} and the output layer are connected deterministically through snapshots of the basis functions $\{\phi_{m}^{(O)}(t)\}_{m=1}^{M^{(O)}}$. The proposed decoder is essentially and functionally consistent with NNBR, a neural network designed for scalar input and functional output, developed by \citet{wu2022SIFO}, since both approaches decompress the scalar-valued basis coefficients to the functional curves in a linear manner, ensuring the use of backpropagation in model training.

Likewise, an advantage of this layer architecture is that it easily handles irregularly-spaced as explained in \citet{wu2022SIFO} and provides us with a smooth reconstruction of the input functional data, one that can be evaluated at any point on the domain thus effectively smoothing the functional data while simultaneously learning a meaningful and useful representation.

\subsubsection{Training the proposed FAE}
\begin{figure}
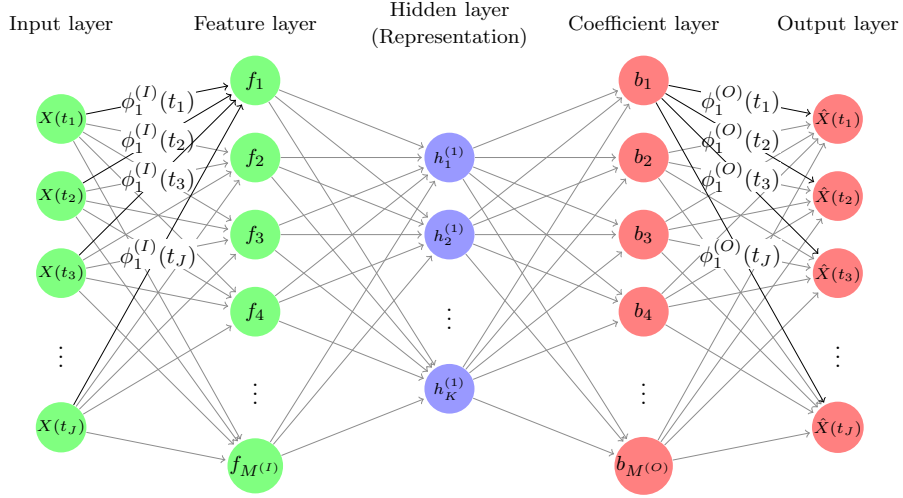

\centering
\resizebox{1\textwidth}{!}{
\begin{neuralnetwork}[nodespacing=12mm,
layerspacing=30mm, maintitleheight=1em, layertitleheight=11em, height=1, toprow=false, nodesize=22pt, style={},title={}, titlestyle={}]
    \newcommand{\x}[2]{\footnotesize \ifnum #2=5 $X(t_{J})$ \else $X(t_#2)$ \fi}
    \newcommand{\f}[2]{\small \ifnum #2=6 $f_{M^{(I)}}$ \else $f_#2$ \fi}
    \newcommand{\fhat}[2]{\small \ifnum #2=6 $b_{M^{(O)}}$ \else $b_#2$ \fi}
    \newcommand{\xhat}[2]{\footnotesize \ifnum #2=5 $\hat{X}(t_{J})$ \else $\hat{X}(t_#2)$ \fi}
    \newcommand{\hfirst}[2]{\footnotesize \ifnum #2=4 $h^{(1)}_{K}$ \else $h^{(1)}_#2$ \fi}
    \newcommand{\hsecond}[2]{\footnotesize $h^{(2)}_#2$}

    \inputlayer[count=5, exclude={4}, bias=false, text=\x, title = {\small Input layer}]
    \inputlayer[count=6, exclude={5}, bias=false, text=\f, title = {\small Feature layer}] 
    \foreach \n in {1,2,3,5}{
       \foreach \m in {2, 3, 4, 6}{
        \link[style={}, labelpos=midway, from layer=0, from node=\n, to layer=1, to node=\m]
        }
    }
    \newcommand{\wfirst}[4]{\ifnum #2=5 $\phi_{#4}^{(I)}(t_{J})$ \else $\phi_{#4}^{(I)}(t_#2)$ \fi}
    \setdefaultlinklabel{\wfirst}
    \foreach \n in {1,2,3,5}{
        \foreach \m in {1}{
            \link[style={}, labelpos=midway, from layer=0, from node=\n, to layer=1, to node=\m, style=black]
        }
    }

    \hiddenlayer[count=4, bias=false, exclude={3}, text=\hfirst, title={\small Hidden layer \\ (Representation)}]
    \newcommand{\wnone}[4]{}
    \setdefaultlinklabel{\wnone}
    \foreach \n in {1,...,4,6}{
       \foreach \m in {1,2,4}{
        \link[style={}, labelpos=midway, from layer=1, from node=\n, to layer=2, to node=\m]
        }
    }

    \outputlayer[count=6, bias=false,exclude={5}, text=\fhat, title={\small Coefficient layer}]
    \foreach \n in {1,2,4}{
       \foreach \m in {1,...,4,6}{
        \link[style={}, labelpos=midway, from layer=2, from node=\n, to layer=3, to node=\m]
        }
    }
    
    \outputlayer[count=5, exclude={4}, text=\xhat, title ={\small Output layer}] 
    \foreach \n in {2,...,4,6}{
       \foreach \m in {1,...,3,5}{
        \link[style={}, labelpos=midway, from layer=3, from node=\n, to layer=4, to node=\m]
        }
    }
    \newcommand{\wsecond}[4]{\ifnum #4=5 $\phi_{#2}^{(O)}(t_{J})$ \else $\phi_{#2}^{(O)}(t_#4)$ \fi}
    \setdefaultlinklabel{\wsecond}
    \foreach \n in {1}{
        \foreach \m in {1,...,3,5}{
            \link[style={}, labelpos=midway, from layer=3, from node=\n, to layer=4, to node=\m, style=black]
        }
    }
    \path (L0-3) -- node{$\vdots$} (L0-5);
    \path (L1-4) -- node{$\vdots$} (L1-6);
    \path (L2-2) -- node{$\vdots$} (L2-4);
    \path (L3-4) -- node{$\vdots$} (L3-6);
    \path (L4-3) -- node{$\vdots$} (L4-5);
\end{neuralnetwork}}
\vspace{0em}
\caption{A graphical representation of the FAE we propose for discrete functional data. The model represented only has a single hidden layer $h$, that serves the role of latent representation.}
\label{fig_DFAE}
\end{figure}

A full architecture (with $L = 1$) of the proposed FAE is displayed in Figure \ref{fig_DFAE}. As detailed in section \ref{sec_encoder_w/_feature_layer} and section \ref{sec_decoder_w/_coef_layer}, a deterministic \textit{feature layer} of size $M^{(I)}$ is created to follow the input layer without using any unknown parameters or weights for neuron calculation, and each neuron in the \textit{feature layer} produces a scalar value computed as the numerical approximation of the inner product of the input $X(t_j)$ and the preselected basis function $\phi_{m}^{(I)}(t_j)$ over the observed timestamp. On the other end, a \textit{coefficient layer} of $M^{(O)}$ scalar-valued neurons is handcrafted as the second to last layer, and it connects the output layer through the known basis functions $\phi^{(O)}_{m}(t)$, making the output layer also deterministic. Layers between the \textit{feature layer} and the \textit{coefficient layer} share the same structure as a conventional autoencoder. This specific structure is the essence of the proposed FAE.

Same as traditional autoencoders, the training process of FAEs comprises of two components, the \textit{forward propagation} and the \textit{backward propagation}, and can be operated using existing neural network libraries or modules, such as \texttt{pytorch} \citep{pytorch} and \texttt{tensorflow} \citep{tensorflow}. The \textit{forward propagation} has been previously depicted and is summarized in algorithm \ref{Algorithm_DFAE_forward}, here we put emphasize on the \textit{backward propagation} that updates the network parameters using gradient-based optimizers.

\begin{algorithm}
\newcommand{\hrulealg}[0]{\vspace{2mm} \hrule \vspace{1mm}}
\linespread{1.1}\selectfont
\caption{FAE Forward Pass}
\label{Algorithm_DFAE_forward}
\KwIn{$\boldsymbol{X} = \{X(t_1), X(t_2), ..., X(t_J)\}$}
\KwOut{$\boldsymbol{\hat{X}} = \{\hat{X}(t_1),\hat{X}(t_2),...,\hat{X}(t_J) \}$}
\textbf{Hyper-parameters:}$\{\phi^{(I)}_{m}(t_j)\}_{m=1}^{M^{(I)}}, \{\phi^{(O)}_{m}(t_j)\}_{m=1}^{M^{(O)}}$, $\omega_{j}$ for all $j$, a pre-defined network NN($\theta$) with $L$ hidden layers, $K^{(l)}$ neurons in the $l$-th hidden layer, activation functions $g_1, ..., g_{L}$, $E$ epochs, Optimizer (including learning rate $\varrho$), etc.
\hrulealg
\nl \textbf{Input Layer} $\rightarrow$ \textbf{\textit{Feature Layer}}\\
$\{X(t_j)\}_{j=1}^{J} \rightarrow f_{m} =\sum_{j=1}^{J}\omega_{j}X(t_{j})\phi^{(I)}_{m}(t_{j}), m \in \{1, 2, ..., M^{(I)}\}$

\nl \textbf{\textit{Feature Layer}} $\rightarrow$ \textbf{\textit{Coefficient Layer}}\\
 $\{f_{m}\}_{m=1}^{M^{(I)}} \rightarrow b_{m} = \sum_{k=1}^{K^{L}}c_{mk}^{(O)}g_{L}\left(\cdots g_{1}\left( \sum_{m=1}^{M^{(I)}}c_{mk}^{(I)}f_{m} \right)\right),  m \in \{1, 2, ..., M^{(O)}\}$\\
 Specifically, the $k$-th neuron in the $l$-th hidden layer is constructed the same way as that in conventional neural networks as $h_{k}^{(l)} = g_{l}(\sum_{k=1}^{K^{(l)}}h_{k}^{(l-1)}w^{(l)})$, and $w^{(l)}$ are the scalar network weights.
 
\nl \textbf{\textit{Coefficient Layer}} $\rightarrow$ \textbf{Output Layer}\\
$\{b_{m}\}_{m=1}^{M^{(O)}} \rightarrow  \hat{X}(t_{j}) = \sum_{m=1}^{M^{(O)}}b_{m}\phi^{(O)}_{m}(t_{j}),  j \in \{1, ..., J\} $

\textbf{return} $ \{\hat{X}(t_1),\hat{X}(t_2),...,\hat{X}(t_J) \}$
\end{algorithm}

Let $\theta = \{c^{((I)}_{mk}, c^{(O)}_{mk}, \eta\}$ denote the collection of network parameters, where $\eta$ stands for all the network weights involved in connecting the hidden layers. The training process targets at finding $\hat{\theta} = \operatorname*{argmin}_{\theta} L(X(t_{j}), \hat{X}(t_{j}))$, and we employ the standard mean-square-error (MSE) between $X(t_{j})$ and $\hat{X}(t_{j})$ across all the observed time points $j$ and subjects $i$ in the training set as the reconstruction error of the FAE, in specific, $L(X(t_{j}), \hat{X}(t_{j})) = \frac{1}{N_{\text{train}}} \sum_{i=1}^{N_{\text{train}}}\sum_{j=1}^{J} (X_i(t_j)-\hat{X}_i(t_j))^2$. We design the output layer of FAE to be a linear combination of some preselected basis functions $\{\phi_{m}^{(O)}\}_{m=1}^{M^{(O)}}$ and the neurons $\{b_m\}_{m=1}^{M^{(O)}}$ outputted by the second to last layer (the \textit{coefficient layer}), and therefore the neuron $\hat{X}(t_j)$ in the output layer of FAE, which is the snapshot of the reconstructed curve $\hat{X}(t)$ at time $t_j$ is the vector product of $\{b_m\}_{m=1}^{M^{(O)}}$ and $\{\phi_{m}^{(O)}\}_{m=1}^{M^{(O)}}$ evaluated at the specific $t_j$. The linear relation between the \textit{coefficient layer} and the output layer, together with the differentials of the known basis functions, ensure the feasibility of computing the gradient of $(X_i(t_j)-\hat{X}_i(t_j))^2$ with respect to the coefficients $b_m$ as:
\begin{align}
    \frac{\partial L}{\partial b_{m}} =             \frac{\partial L}{\partial \hat{X}(t_j)}\frac{\partial \hat{X}(t_j)}{\partial b_{m}}.
\end{align}

The gradient with respect to the network weights $\eta$ in the remaining layers prior to the \textit{coefficient layer} can be subsequently computed in the backward manner as that in a classic neural network until reaching the \textit{feature layer}, while no any further gradient calculation is made from the \textit{feature layer} back to the input layer because they are connected by the predefined input basis functions $\{\phi_{m}^{(I)}\}_{m=1}^{M^{(I)}}$, instead of some network parameters in need of training. Algorithm \ref{Algorithm_DFAE_backward} details the gradients calculating procedure used to update network parameters in the \textit{backpropagation}.

\begin{algorithm}[h]
\newcommand{\hrulealg}[0]{\vspace{2mm} \hrule \vspace{1mm}}
\linespread{1.1}\selectfont
\caption{FAE Backward Pass}
\label{Algorithm_DFAE_backward}
\KwIn{$\theta_{\text{current}}, \{X(t_1), X(t_2), ..., X(t_J)\}, \{\hat{X}(t_1),\hat{X}(t_2),...,\hat{X}(t_J) \}$}
\KwOut{$\theta_{\text{updated}}$}
\textbf{Hyper-parameters:}$\{\phi^{(I)}_{m}(t_j)\}_{m=1}^{M^{(I)}}, \{\phi^{(O)}_{m}(t_j)\}_{m=1}^{M^{(O)}}$, $\omega_{j}$ for all $j$, a pre-defined network NN($\theta$) with $L$ hidden layers, $K^{(l)}$ neurons in the $l$-th hidden layer, activation functions $g_1, ..., g_{L}$, $E$ epochs, Optimizer (including learning rate $\varrho$), etc.
\hrulealg
\nl Compute loss function $L(X(t_{j}), \hat{X}(t_{j}))$

\nl Set $\theta = \theta_{\text{current}}$

\nl \textbf{Output Layer} $\rightarrow$ \textbf{\textit{Coefficient Layer}}\\
 $\frac{\partial  L(\theta)}{\partial b_{m}} =  \frac{\partial  L(\theta)}{\partial \hat{X}(t_j)}\frac{\partial \hat{X}(t_j)}{\partial b_{m}}$, because $\hat{X}(t_j) = f(b_{m})$ and $f'(b_{m})$ exists
 
\nl \textbf{\textit{Coefficient Layer}} $\rightarrow$ \textbf{\textit{Feature Layer}}\\
$\frac{\partial L(\theta)}{\partial \theta}$, same gradient calculation as used in traditional neural networks.

\nl \textbf{\textit{Feature Layer}} $\rightarrow$ \textbf{Input Layer}\\
No gradient calculation involved (deterministic operation)

\nl Update NN parameters $\theta^{\ast}$

\textbf{return} $\theta_{\text{updated}} = \theta^{\ast}$
\end{algorithm}


\subsection{FAE as a functional data smoother}
By design, our proposed FAE outputs a smooth continuous curve over the entire interval of interest as an estimate of the underlying generative stochastic process for any input distinctly observed functional data by
\begin{align}
\hat{X}(t) 
&=\sum_{k=1}^{K^{(L)}}\sum_{m=1}^{M^{(O)}} h_{k}^{(L)}c^{(O)}_{mk}\phi^{(O)}_{m}(t) = \sum_{m=1}^{M^{(O)}}b_{m}\phi^{(O)}_{m}(t),
\end{align}
which is achievable thanks to the continuity of the preselected basis functions $\phi^{(O)}_{m}(t)$'s. This is a core design choice made so that the FAE we propose acts not only as a representation learner but a smoother itself and could substitute other smoothing processes such as fitting a B-Spline model. 

Following the tradition in FDA, we can promote the smoothness of the output curves by adding a roughness penalty to the objective function of the FAE. With a consideration for computational simplicity, among different choices of roughness penalties, we propose to apply the difference penalty on the elements of the \textit{coefficient layer} as they act as the basis coefficients in the output functional curves. Consequently, including such a penalty term leads to the following objective function,
\begin{align}
    L_{\boldsymbol{pen}} = \frac{1}{N_{\text{train}}}
    \sum_{i=1}^{N_{\text{train}}}\left(\sum_{j=1}^{J}\left(X_{i}(t_{j})-\hat{X}_{i}(t_{j})\right)^2+\lambda\sum_{m=3}^{M^{(O)}}(\Delta^{2} b_{im})^{2}\right)
    \label{penalized_loss.coef}
\end{align}
and $\Delta^{2}b_{m} = b_{m}-2b_{m-1}+b_{m-2}$, where $b_{im}$ is the $m$-th neuron in the \textit{coefficient layer} for the $i$-th subject, and parameter $\lambda$ controls the smoothness. In implementation, we suggest applying roughness penalty when the $M^{(O)}$ is relatively large ($M^{(O)} >> J$) and the optimal $\lambda$ can be selected using cross validation.

\subsection{FAE for irregularly-spaced observations}

For many existing FDA models, it is quite common to assume that the observed discrete functional data are regularly spaced. A benefit of our designed FAE is that it is free of this assumption and its input layer can actually be of flexible size because of the proposed \textit{feature layer} applied in the early stages of the model. 

As detailed in section \ref{sec_encoder_w/_feature_layer}, we express the input functional weights $w^{(I)}_{k}(t_j)$ by a fixed representation of $\sum_{m=1}^{M_{k}^{(I)}}c^{(I)}_{mk}\phi^{(I)}_{mk}(t_j)$, and therefore every discrete functional input $X_{i}(t_{ij}), j=1, ..., J_{i}$, where $J_{i}$ varies with $i$, are all equivalently projected to the same basis functions $M^{(I)}$, forming the $t_{ij}$-free features $f_{im} =\sum_{j=1}^{J_{i}}\omega_{ij}X_{i}(t_{ij})\phi^{(I)}_{m}(t_{ij}), m = \{1, 2, ..., M^{(I)}\}$. These $M^{(I)}$ features then participate in the following forward pass in place of the actual functional inputs $X_{i}(t_{ij})$ for training the same set of network parameters including the input weight coefficients $c^{(I)}_{mk}$, which are free of $i$.

The designed \textit{feature layer}, coupled with the input functional weight representation, digests the irregular inputs by generalizing the problems of estimating irregular snapshots of input weight functions to estimating input weight coefficients that are consistent over all subjects.  

\section{Connection with existing models}
\label{sec:connection}
\subsection{Relation with FPCA}

As previously pointed out by \citet{BaldiPierre1989Nnap}, \citet{BengioY2013RLAR}, and \citet{Bourlard1988-fk}, a single-hidden-layer linear autoencoder with its objective function being the squared reconstruction error, i.e., $L = \sum_{i=1}^{N_{\text{train}}} \left\|X_i - \hat{X}_i \right\|^{2} = \sum_{i=1}^{N_{\text{train}}} \left \|X_i - W_{\text{d}}W_{\text{e}} X_i \right \|^{2} = \sum_{i=1}^{N_{\text{train}}}\sum_{p=1}^{P} \left \{X_{ip} - (W_{\text{d}}W_{\text{e}}X_i)_p  \right \}^{2}$,
where $X_i = \{X_{i1}, X_{i2}, ..., X_{iP}\}$, $W_{\text{d}}$, $W_{\text{e}}$ denote the $i$-th network input of $P$ dimensions, weight matrix of the decoder and weight matrix of the encoder, respectively, is approximately identical to the conventional PCA, because such an autoencoder is learning the same subspace as the PCA. More precisely, the unique global minimum of $L$ is corresponding to the orthogonal projection of $X$ onto the subspace spanned by the leading eigenvectors (principal components) of the covariance matrix of $X$. It is worth mentioning that at the global minimum, the uniqueness occurs with the global map $W_{\text{d}} \times W_{\text{e}}$, 
while the matrices $W_{\text{d}}$ and $W_{\text{e}}$ may not be unique. This is because, for multiple appropriate $C$ we have $W_{\text{d}} \times W_{\text{e}} = (W_{\text{d}}C)(C^{-1}W_{\text{e}})$. In other words, the mapping $W_{\text{d}} \times W_{\text{e}}$ is unique but not the encoder and decoder weight matrices.

 When it comes to the functional scenario, a homogeneous relationship exists between FAE and FPCA. For a single-hidden-layer FAE with linear activation function under continuous functional data setting, the objective function measuring the mean squared reconstruction error turns out to be
\begin{align}
    L & = \sum_{i=1}^{N_{\text{train}}} \left\|X_i - \hat{X}_i \right\|^{2}
    =\sum_{i=1}^{N_{\text{train}}} \int_{\mathcal{T}}\left\{X_i(t) - \sum_{k=1}^{K^{(1)}}\left(\int_{\mathcal{T}}X_i(t)w^{(I)}_k(t)dt\right) w^{(O)}_k(t)\right\}^{2}dt.
    \label{Loss_FAE_continuous}
\end{align}

\citet{fda} concluded that the aforementioned fitting criterion is minimized when the orthonormal-restricted weight functions $w^{(\cdot)}(t)$ are precisely the same set of principal component weight functions of the functional data $X(t)$. Hence, training a one-hidden-layer linear FAE with respect to the squared reconstruction error criterion and an orthonormal constrain on functional weights is exactly approaching to project the input $X(t)$ onto the subspace generated by the first functional principal components, the same space learned by FPCA. 

For discrete functional data, the objective model becomes
\begin{align}
    L & = \sum_{i=1}^{N_{\text{train}}} \left\|X_i - \hat{X}_i \right\|^{2} = \sum_{i=1}^{N_{\text{train}}} \frac{1}{J}\sum_{j=1}^{J}\left\{X_i(t_j) - \sum_{k=1}^{K^{(1)}}\left(\sum_{j=1}^{J}\omega_jX_i(t_j)w^{(I)}_k(t_j)\right) w^{(O)}_k(t_j)\right\}^{2}.
    \label{Loss_FAE_discrete}
\end{align}
It is important to notice that this approximation can lead to some difference which should progressively decreases as $J$ increases. Consequently, for relatively large values of $J$, the FAE optimized by minimizing Eq.(\ref{Loss_FAE_discrete}) will yield functional weights that are approximately the same as those obtained by minimizing Eq.(\ref{Loss_FAE_continuous}). To put it differently, when subjected to the orthonormal constraint on the functional weights, the FAE that minimizes the objective model Eq.(\ref{Loss_FAE_continuous}) is effectively learning the empirical projection of $X(t)$ onto the same space as FPCA does. Importantly, the proposed FAE with discrete configuration generalizes FPCA up to a few approximations, and the functional weights produced by FAE can be identically interpreted as the FPCs in FAE.

\subsection{Relation with AE}
As pointed out in section \ref{sec_DFAE}, the proposed FAE is structurally similar to the classic AEs based on fully connected neural networks. The main difference lies in the first and last layers. In detail, a classic AE consists of network weights (and bias) free of restrictions and the training task aims at optimizing these vectors of network parameters. The FAE we developed also includes such weights to link layers between the \textit{feature layer} and \textit{coefficient layer}, however, the difference lies in the deterministic weights before the \textit{feature layer} and after the \textit{coefficient layer}, which are comprised of snapshots of continuous basis functions. With the goal of optimizing the non-deterministic weights, the training process of FAE follows the same rule as used in classic AE. The FAE can be regarded as an extension of AE with some deterministic layers added to both ends of the network.

The addition of \textit{feature layer} to the conventional AE architecture enables the FAE to quickly summarize the underlying temporal relationship among observed time span into neurons that actually step into the network, resulting in the faster convergence and better generalization in network training, compared to that of conventional AE. Meanwhile, thanks to the application of the functional output weights, the FAE we developed can recover the discrete functional data to smooth curves over a continuous interval, satisfying the smoothness requirement of functional data, while the classic AE is limited to output discontinuous functions evaluated at some discrete timestamp of observations. Additionally, our method is capable to efficiently handle irregularly spaced functional input along with its underlying correlation in the \textit{feature layer} by adjusting the weights $\omega$ used for numerical integration calculation, while AE has to take care of the shortage of observations at some time points by training the model with some null-valued input for the corresponding neurons in the input layer. The designed structure benefits our method with better performances in less computational cost when manipulating irregularity, which is further highlighted by a series of simulation studies in the following section.

\section{Simulation study}
\label{sec:simulation}
In this section, we aim to compare our proposed FAE with the two existing baseline methods it extends, FPCA and AE respectively, for representation learning and curve smoothing from discretely observed functional data. We concentrate on investigating the effectiveness of our method compared to FPCA in capturing the potential nonlinear relationship as well as evaluating the smoothing ability and computational efficiency of the FAE compared to AE.

\subsection{Simulation setup}

\subsubsection{Data generation}
We generate the data by first sampling a $d$-dimensional representation $\boldsymbol{Z}$ from a Gaussian mixture model. The mean vector and the covariance matrix of each component are designed so that components are separable. We then apply a function $f(\cdot)$ that maps the representations $\boldsymbol{Z}$, to a set of $M$-dimensional basis coefficients $B_{m}$. Finally, we produce the continuous functional data using a linear combination of $M$ basis functions $\psi_m(t)$'s and the basis coefficients $B_{m}$:
\begin{align}
    X(t) = \sum_{m=1}^{M}B_{m}\psi_m(t) =  f(\boldsymbol{Z})\boldsymbol{\psi}.
\end{align}
Finally, we evaluate $X(t)$ at some discrete times $\{t_1, t_2, ..., t_J\} \in [0, 1]$ to obtain a discrete version of the functional data. 

The basis system used $\psi_m(t)$ are B-spline basis system with an order of 4, and the number of bases $M$ varies throughout experiments. In terms of the mapping function $f(\cdot)$, we employ a neural network $\text{NN}(\cdot)$ with multiple architectures aiming to create different mapping paths from the representation vector $\boldsymbol{Z}$ to the basis coefficients of the functional data. The neural network takes the $d$-dimensional representation vector as input and outputs the $M$-dimensional basis coefficients. We apply a simple model with no hidden layers and linear activation function for linear scenarios, and neural networks with at least one hidden layer and nonlinear activation functions for nonlinear scenarios. 

An optional Gaussian noise can be further added to the discrete functional curve to mimic observational errors. The component the representation was sampled from is used as the label for the functional data in classification experiments. 

\subsubsection{Implementation of models}

\textbf{FPCA} linearly encodes functional curves to FPC scores $\xi_{im}$'s with corresponding FPCs $\psi_{m}(t)$'s. We implement FPCA in \texttt{python} relying on the \textbf{scikit-fda} library \citep{skfda}. The discrete functional data are firstly converted to smooth functions using basis expansion with customized number of B-spline basis functions, and then the conventional FPCA is performed on the estimated curve with a user-defined number of FPC. The resulting FPC scores serve as the scalar representation of the functional data and are used for further statistical analyses. 

\textbf{AE} based on densely feed-forward-network architecture can learn an encoding from the functional trajectory observed at discrete time points to a lower-dimensional vector of representation without considering any temporal correlation among the discrete observations. We design the input layer of AE to have $J$ neurons with the $j$-th neuron representing the snapshot of the functional curve observed at $t_j$. We adopt different architecture with a bottleneck hidden layer being the layer producing representation and attempted with both linear and nonlinear activation functions and initialize the network weights to random values drawn from $\mathcal{N}(0, \sigma)$. We implement the AE using the \textbf{pytorch} library.

Lastly, we implement the proposed \textbf{FAE} using the \textbf{pytorch}, coupled with the \textbf{scikit-fda} library for applying the basis expansion to functional weights. Analogously, we attempt with different architecture with a hidden layer for extracted representation, employ both linear and nonlinear activation functions in model training, and initialize weights randomly by sampling them from a Gaussian distribution $\mathcal{N}(0, \sigma)$.

\subsection{Results}
Series of simulations are performed under various scenarios to investigate the performance of the proposed method in both prediction and classification against those of FPCA and AE, separately. The prediction error is measured by the mean squared prediction error (MSE$_{\text{p}}$) averaged across the number of samples and the number of observed time points in the test set, while the classification accuracy, P$_{\text{classification}}$, is calculated as the percentage of test observations that can be labelled correctly by a logistic regression based on the representations extracted. For each scenario, we report the mean and standard deviation (SD) of the evaluation metrics across all replications.

\subsubsection{FAE vs. FPCA}
\label{sec: FAE_vs_FPCA}
\textbf{\textit{Scenario 1.1 (Linear \& Regular)}}: 6000 discrete functional observations evaluated at 21 equally spaced points over the interval $[0,1]$ are simulated. A five-dimensional Gaussian mixture model with three components is used to generate the representations and the resulting functional curves are labelled with class 1, 2 and 3. A neural network without hidden layers and a linear activation function is performed to map the representation to the basis coefficients. We employ 8 B-spline basis functions ($M=8$) along with the aforementioned basis coefficients to express the underlying functional curves. 

We assign 80\% of the observations by random to the training set and the remainder to the test set. The FPCA and two types of FAE are successively trained and the model details are summarized in Table S1 in the supplementary document.

\begin{table}
    \centering
    \caption{Means and standard deviations (displayed inside parentheses) of prediction error and classification accuracy of functional autoencoder with Identity activation function (FAE(Identity)), functional autoencoder with Softplus activation function (FAE(Softplus)) and functional principal component analysis (FPCA) on 10 random test data sets in Scenario 1.1, with the best results being highlighted in bold.}
    \begin{tabular}{@{} c | c | c c c@{}}
    \toprule
    \multicolumn{2}{c|}{} & \thead{\textbf{FAE} \\ (Identity)} & \thead{\textbf{FAE} \\  (Softplus)} &  \thead{\textbf{FPCA}} \\
     \midrule  
     \multirow{3}{*}{ \thead{MSE$_{\text{p}}$}}& 3 Reps &  0.0050(0.0001) & \textbf{0.0045}(0.0005) & 0.0052(0.0001)\\
     & 5 Reps  &  \textbf{0.0019($<$0.0001)} & 0.0022(0.0003) & 0.0021($<$0.0001)\\
     & 10 Reps & \textbf{0.0009($<$0.0001)} & 0.0017(0.0005) & 0.0010($<$0.0001)\\
     \midrule  
     \multirow{3}{*}{ \thead{P$_{\text{classification}}$}} & 3 Reps &  87.24\%(0.93\%) & \textbf{87.72\%}(1.62\%) & 87.68\%(0.78\%)\\
    & 5 Reps & 87.94\%(0.81\%) & 86.53\%(0.94\%) & \textbf{89.21\%}(0.78\%)\\
    & 10 Reps  & 89.16\%(0.75\%) & \textbf{89.61\%}(0.99\%) & 89.22\%(0.70\%)\\
   \bottomrule 
    \end{tabular}
    \label{tab:DFAE_FPCA_sce1.1}
\end{table}

\textbf{\textit{Scenario 1.2 (Nonlinear \& Regular)}}: We generate 3000 functional observations discretely measured at 51 equally spaced points over the interval $\mathcal{T} = [0,1]$. We sample a 5-dimensional representations for each curve from a 3-component Gaussian mixture model and label the associated functional curves with class 1, 2 and 3. We map the representations to the basis coefficients using a neural network with one hidden layer with 20 neurons and Sigmoid activation function. Afterwards, individual functional curve is constructed using 10 B-spline basis functions and the basis coefficients described above. 

We continue to randomly generate training and test sets that contain 80\% and 20\% observations respectively. Again, we put FPCA, linear FAE and nonlinear FAE in comparison with model configuration adjusted and detailed in Table S2 in the supplementary document.

\begin{table}
    \centering
    \caption{Means and standard deviations (displayed inside parentheses) of prediction error and classification accuracy of functional autoencoder with Identity activation function (FAE(Identity)), functional autoencoder with Sigmoid activation function (FAE(Sigmoid)) and functional principal component analysis (FPCA) on 10 random test data sets in Scenario 1.2, with the best results being highlighted in bold.}
    \begin{tabular}{@{} c| c | c c c@{}}
    \toprule
     \multicolumn{2}{c|}{} & \thead{\textbf{FAE} \\ (Identity)} & \thead{\textbf{FAE} \\  (Sigmoid)} &  \thead{\textbf{FPCA}} \\
     \midrule  
     \multirow{3}{*}{MSE$_{\text{p}}$} & 3 Reps &  0.0070(0.0002) & \textbf{0.0038}(0.0002) & 0.0070(0.0002)\\
     & 5 Reps  & 0.0035(0.0001) & \textbf{0.0026}(0.0004) &  0.0036(0.0001)\\
     &  10 Reps & \textbf{0.0013}($<$0.0001) & 0.0014($<$0.0001)  & \textbf{0.0013}($<$0.0001)\\
    \midrule  
      \multirow{3}{*}{P$_{\text{classification}}$} & 3 Reps &  85.05\%(1.08\%) & \textbf{88.68\%}(1.46\%) & 85.17\%(1.06\%)\\
   &  5 Reps & 86.62\%(1.06\%) & \textbf{92.42\%}(1.02\%) &  86.65\%(1.28\%)\\
   &  10 Reps & 87.55\%(1.13\%) & \textbf{91.20\%}(1.06\%) & 87.53\%(1.26\%)\\
   \bottomrule 
    \end{tabular}
    \label{tab:DFAE_FPCA_sce1.2}
\end{table}

Table \ref{tab:DFAE_FPCA_sce1.1} and Table \ref{tab:DFAE_FPCA_sce1.2} summarize the predictive and classification performances of the proposed FAE and FPCA. In the linear \& regular context, we observe that all three approaches in comparison yield similar performances in both prediction and classification for most representation attempts.

In contrast, under the nonlinear scenario, both linear FAE (FAE with Identity activation function) and FPCA generate relatively higher $\text{MSE}_{\text{p}}$ and lower $\text{P}_{\text{classification}}$ due to the violation of the linearity assumption. Meanwhile, the FAE with Sigmoid activation function retains superior performances on both prediction and classification in comparison to the other linear approaches, with only minimal difference when predicting with 10 representations, indicating that the nonlinear FAE can more efficiently and accurately capture and comprise the information carried by the discrete data. 

\begin{figure}
    \includegraphics[width = 1\columnwidth]{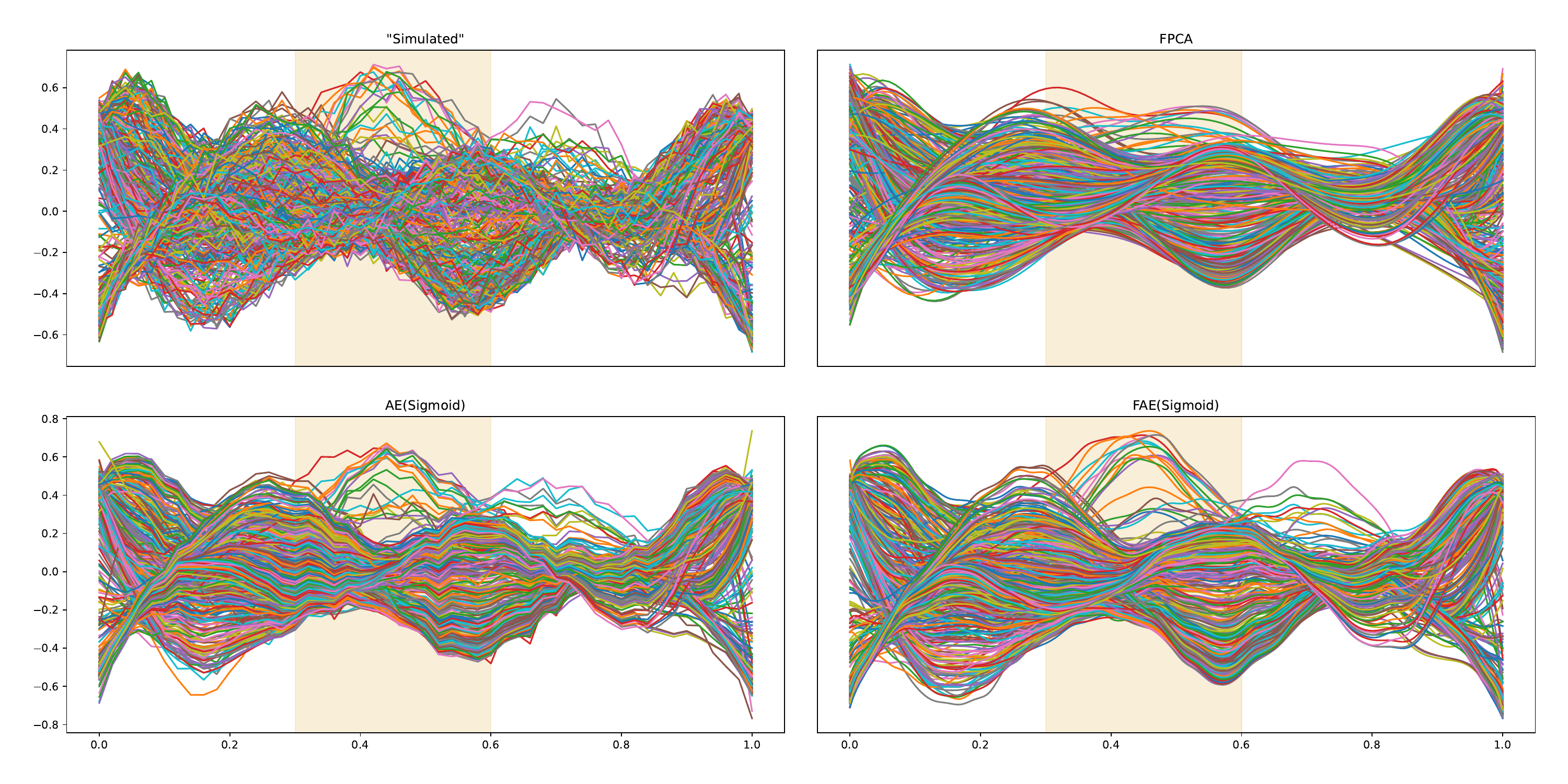}
    \caption{The simulated curves and the curves recovered by  functional principal component analysis (FPCA), classic autoencoder with Sigmoid activation function (AE(Sigmoid)) and functional autoencoder with Sigmoid activation function (FAE(Sigmoid)) using 5 representations for a random test set in Scenario 1.2 and Scenario 2.1.}
    \label{fig:sce1.2+sce2.1_Recovery}
\end{figure}

With regard to curve smoothing, as displayed in Figure \ref{fig:sce1.2+sce2.1_Recovery}, both FPCA and FAE produce smooth curves based on the inputted discrete observations, while the designed FAE demonstrates additional benefits in curve recovery. Plainly, FAE can not only correctly reconstruct the complete moving trend but also sensitively capture the individual pop-up variations, e.g. the local $\cap$-shaped mode appearing in the shaded interval.

\subsubsection{FAE vs. AE}
\label{sec: FAE_vs_AE}
\textbf{\textit{Scenario 2.1 (Nonlinear \& Regular)}}: The simulated data used in the scenario 1.2 in section \ref{sec: FAE_vs_FPCA} is simultaneously applied for a comparison between FAE and AE. Again, 80\% of the random observations are assigned to the training set and the remaining 20\% to the test set. Given this scenario follows the nonlinear setting, we put emphasize on the nonlinear models by training the baseline model AE and the proposed FAE with model configurations listed in Table S3 in the supplementary document. 

\begin{sidewaystable}
    \centering
    \caption{Means and standard deviations (displayed inside parentheses) of prediction error and classification accuracy of functional autoencoder with Sigmoid activation function (FAE(Sigmoid)) and classic autoencoder with Sigmoid activation function (AE(Sigmoid)) on 10 random test data sets in Scenario 2.1, with the better results being highlighted in bold.}
    \begin{tabular}{@{}c| c c c | c c c@{}}
    \toprule
     & \multicolumn{3}{c|}{\textbf{FAE (Sigmoid)}} & \multicolumn{3}{c}{\textbf{AE (Sigmoid)}} \\
      & 3 Reps & 5 Reps & 10 Reps & 3 Reps & 5 Reps & 10 Reps \\
     \midrule
    \thead{MSE$_{\text{p}}$} & \textbf{0.0038}(0.0002) & \textbf{0.0026}(0.0004) & \textbf{0.0014}($<$0.0001) & 0.0046(0.0005) & 0.0030(0.0005) & 0.0124(0.0069) \\
    \midrule
    \thead{P$_{\text{classification}}$} & 88.68\%(1.46\%)  & 92.42\% (1.02\%) & 91.02\%(1.06\%) & \textbf{89.35\%}(1.39\%) & \textbf{92.75\% }(1.15\%)  & \textbf{92.65\%}(1.81\%) \\
    \botrule
    \end{tabular}
    \label{tab:DFAE_AE_sce2.1_train80}
\end{sidewaystable}

Table \ref{tab:DFAE_AE_sce2.1_train80} presents the means and SDs of $\text{MSE}_{\text{p}}$ and $\text{P}_{\text{classification}}$ over 10 replicates trained by AE and FAE in the nonlinear but regularly-spaced-data scenario. We can observe that the two methods achieved competitive performances in representation learning, with the conventional AE giving better results on classifying the curves and the FAE becoming ahead with smaller predictive errors in reconstructing the functional observations. Figure \ref{fig:sce1.2+sce2.1_Recovery} visualizes the simulated curves, and the full trajectories recovered by FPCA, nonlinear AE and nonlinear FAE, respectively. It is clearly shown that the proposed FAE can directly and accurately output smooth curves using the given discrete observations for the entire domain, while AE is limited to discretely recover the curve at the time points with available observations, indicating that the FAE is capable of efficiently capturing the representative information and simultaneously smoothing the discretely functional observation.

\textbf{\textit{Scenario 2.2 (Nonlinear \& Irregular)}}: In this scenario, we continue to use the data simulated in scenario 1.2 and randomly remove measurements in 25 time points (excluding the start and end time point) for each curve to create irregularly and discretely observed functional data, that is, the resulting functional curve contains 26 irregular observations individually over the domain interval $\mathcal{T}$. 

We experiment with two different training set settings: (i) the training set contains 80\% observations; (ii) the training set contains 20\% data, and focus on a comparison between the nonlinear AE and FAE with configurations provided in Table S4 in the supplementary document to examine their performances in handling nonlinearity and irregularity simultaneously. For those time points without observations (randomly removed), we feed the corresponding neurons in the input layer of AE and FAE with 0 and abandon those neurons when computing the loss. When training FAE, we also adjust the weights $\{\omega_{j}\}_{j=1}^{J_{i}}$ individually for each discrete curve $i$ for a numerical integration over all the observed timestamp.

\begin{sidewaystable}
    \centering
    \caption{Means and standard deviations (displayed inside parentheses) of prediction error and classification accuracy of functional autoencoder with Softplus activation function (FAE(Softplus)) and classic autoencoder with Softplus activation function (AE(Softplus)) on 10 random test data sets  when training with 80\% irregularly observed data in Scenario 2.2, with the better results being highlighted in bold.}
    \begin{tabular}{@{}c |c | c c c | c c c@{}}
   \hline
     \multicolumn{2}{c|}{\multirow{2}{*}{}} &
     \multicolumn{3}{c|}{\textbf{FAE (Softplus)}} & \multicolumn{3}{c}{\textbf{AE (Softplus)}} \\
     \multicolumn{2}{c|}{} & 3 Reps & 5 Reps & 10 Reps & 3 Reps & 5 Reps & 10 Reps \\
     \hline
    \multirow{2}{*}{\thead{MSE$_{\text{p}}$}} & epochs=1000 & 0.0031(0.0003) & 0.0023(0.0002) & 0.0014(0.0002)  & 0.0035(0.0002) & 0.0029(0.0003) & 0.0143(0.0127)  \\ 
    & epochs=2000 & 0.0023(0.0001) & 0.0015($<$0.0001) & 0.0010($<$0.0001) & 0.0034(0.0003) & 0.0044(0.0059) &  0.0103(0.0102)\\
    \hline
    \multirow{2}{*}{\thead{P$_{\text{classification}}$}} & epochs=1000  & 86.57\%(1.08\%) & 87.85\%(2.03\%) & 89.22\%(1.17\%) & 89.85\%(1.32\%) & 91.05\%(0.69\%) & 90.58\%(1.59\%) \\
    & epochs=2000  & 88.67\%(1.22\%) & 90.12\%(1.70\%) & 91.75\%(1.10\%) & 90.68\%(1.30\%) & 91.03\%(1.09\%) & 90.73\%(1.66\%)\\
    \hline
    \end{tabular}
    \label{tab:DFAE_AE_sce2.2_80train}
\end{sidewaystable}

\begin{sidewaystable}
    \centering
    \caption{Means and standard deviations (displayed inside parentheses) of prediction error and classification accuracy of functional autoencoder with Softplus activation function (FAE(Softplus)) and classic autoencoder with Softplus activation function (AE(Softplus)) on 10 random test data sets  when training with 20\% irregularly observed data in Scenario 2.2, with the better results being highlighted in bold.}
    \begin{tabular}{@{}c |c | c c c | c c c@{}}
   \hline
     \multicolumn{2}{c|}{\multirow{2}{*}{}} &
     \multicolumn{3}{c|}{\textbf{FAE (Softplus)}} & \multicolumn{3}{c}{\textbf{AE (Softplus)}} \\
     \multicolumn{2}{c|}{} & 3 Reps & 5 Reps & 10 Reps & 3 Reps & 5 Reps & 10 Reps \\
     \hline
     \multirow{5}{*}{\thead{MSE$_{\text{p}}$}} & epochs=1000 & \textbf{0.0057}(0.0009) & \textbf{0.0041}(0.0009)  & \textbf{0.0039}(0.0026) & 0.0386(0.0152) &  0.0730(0.0237) & 0.4591(0.2692) \\
    & epochs=2000  & \textbf{0.0046}(0.0011) & \textbf{0.0030}(0.0004) & \textbf{0.0025}(0.0007) & 0.0194(0.0094) & 0.0464(0.0266) & 0.3579(0.2449)\\
    & epochs=3000  & \textbf{0.0035}(0.0003) & \textbf{0.0027}(0.0003) & \textbf{0.0019}(0.0004) & 0.0104(0.0027) & 0.0230(0.1578) & 0.1917(0.0934)\\
    & epochs=4000 & \textbf{0.0031}(0.0002) & \textbf{0.0021}($<$0.0001) &  \textbf{0.0029}(0.0039) & 0.0086(0.0012) & 0.0093(0.0037) &  0.0968(0.0632) \\
    & epochs=5000 & \textbf{0.0029}(0.0002) & \textbf{0.0019}(0.0001) & \textbf{0.0015}(0.0004) & 0.0094(0.0010) & 0.0070(0.0015) & 0.0588(0.0434) \\
    \hline
    \multirow{5}{*}{\thead{P$_{\text{classification}}$}} & epochs=1000  & 78.32\%(1.10\%) & \textbf{81.59\%}(2.12\%) & \textbf{82.30\%}(2.84\%) & \textbf{78.71\%}(2.97\%) & 80.48\%(2.30\%) & 64.36\%(5.26\%)\\
    & epochs=2000 & 81.40\%(2.00\%) & 83.86\%(1.17\%) & \textbf{83.50\%}(1.20\%) & \textbf{85.30\%}(0.82\%) & \textbf{86.16\%}(1.39\%) & 80.63\%(6.87\%) \\
    & epochs=3000 & 84.09\%(1.06\%) & 85.75\%(1.24\%) & 85.27\%(1.02\%) & \textbf{86.70\%}(1.11\%) & \textbf{87.50\%}(1.57\%) & \textbf{88.61\%}(1.29\%) \\
    & epochs=4000 & 85.05\%(0.72\%) & 86.69\%(1.26\%) & 87.17\%(1.04\%) & \textbf{87.18\%}(1.27\%) & \textbf{88.03\%}(1.17\%) & \textbf{90.05\%}(0.65\%) \\
    & epochs=5000 & 85.53\%(0.94\%) & 87.63\%(1.26\%) & 88.27\%(1.34\%) & \textbf{87.50\%}(1.25\%) & \textbf{88.23\%}(0.86\%) & \textbf{89.98\%}(0.63\%) \\
     \hline
    \end{tabular}
    \label{tab:DFAE_AE_sce2.2_20train}
\end{sidewaystable}

The performances of prediction and classification of nonlinear AE and nonlinear FAE trained with 80\% and 20\% irregularly-spaced functional data are illustrated in Table \ref{tab:DFAE_AE_sce2.2_80train} and Table \ref{tab:DFAE_AE_sce2.2_20train}, separately, with the performances of both models reported for each thousand epochs. We can see that proposed FAE shows more advantages in speedily learning the representation and accurately capturing the information for both prediction and classification, especially when the training epochs remain small. On the other hand, the classic AE needs to gradually master the mapping path in respect of reconstruction error, while its resulting representations can outperform those by FAE in classification when the training cost increases. The visual comparisons of how the mean $\text{MSE}_{\text{p}}$ and mean $\text{P}_{\text{classification}}$ of FAE and AE changes with the number of training epochs for 80\% and 20\% training sizes, corresponding to Table \ref{tab:DFAE_AE_sce2.2_80train} and Table \ref{tab:DFAE_AE_sce2.2_20train}, are presented in Section S1.2 of the supplementary document. As demonstrated, the computational efficiency of the FAE is robust over different representation numbers, which further confirms that the FAE is able to generalize better and converge faster even with fewer epochs and larger batch size over traditional AE that has similar architecture in the matter of curve reconstruction and unsupervised representation learning for classification.

\begin{figure}[ht]
    \centering
    \includegraphics[width = 0.49\columnwidth]{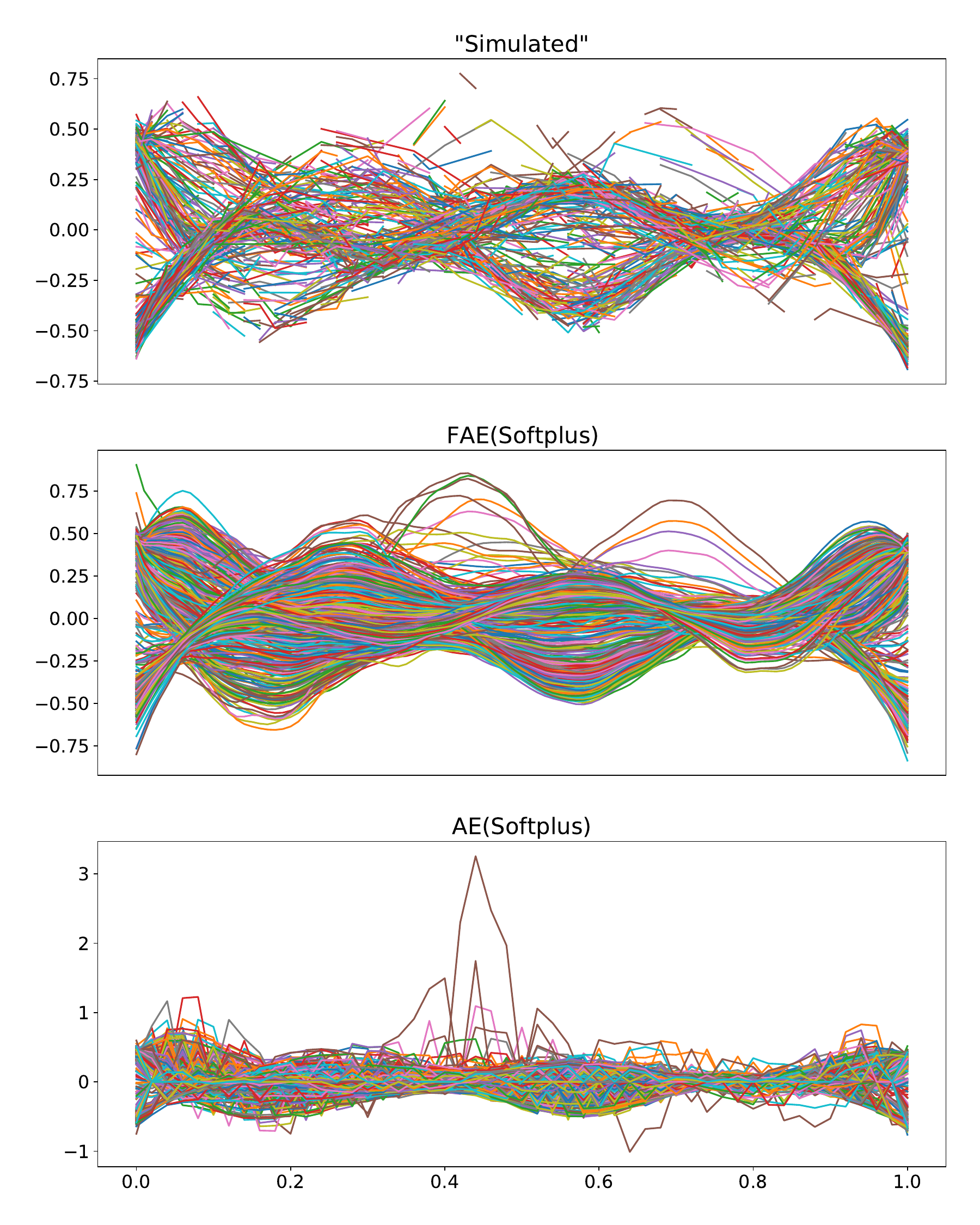}
    \includegraphics[width = 0.49\columnwidth]{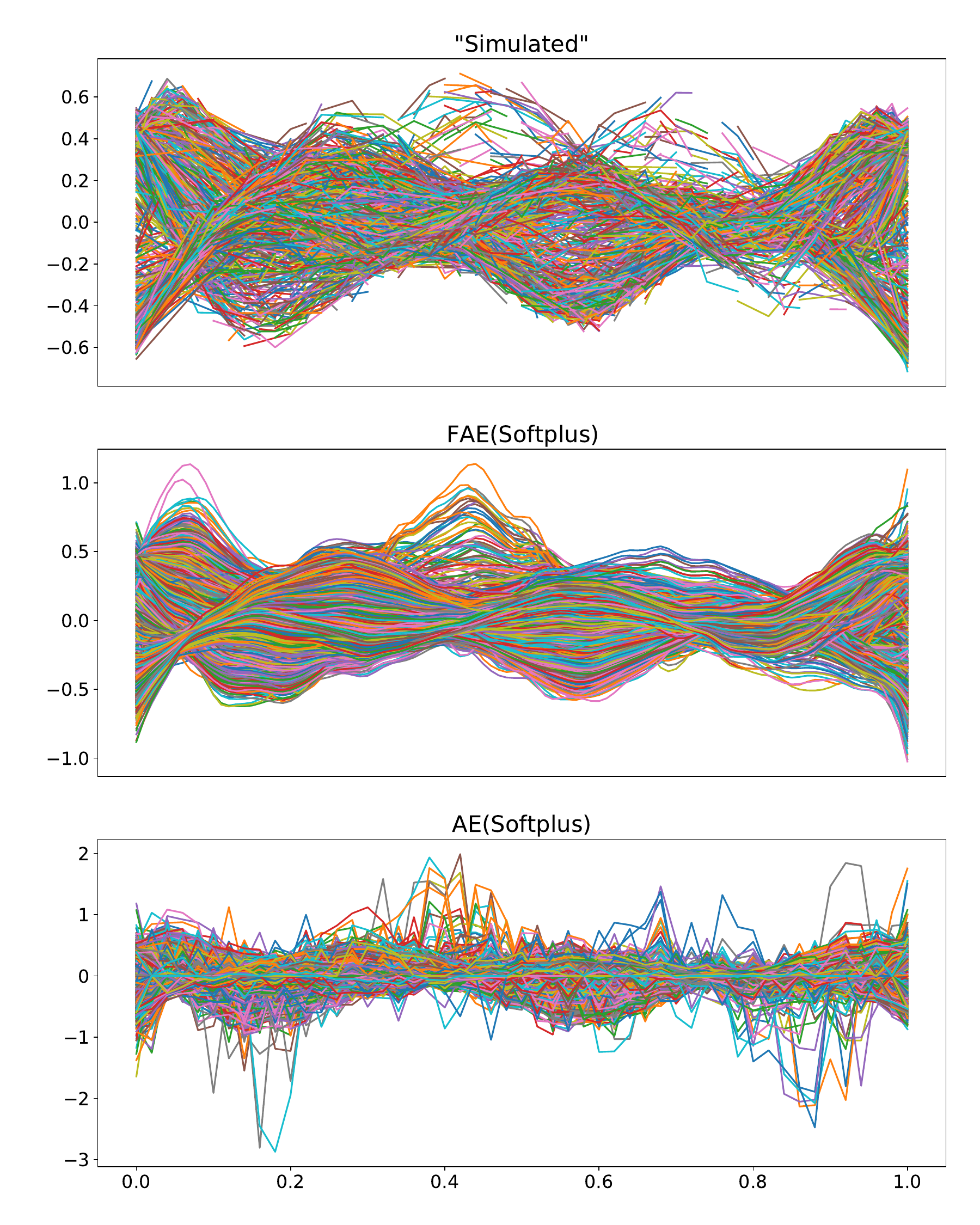}
    \caption{The simulated  irregularly-spaced curves and the curves recovered by classic autoencoder with Softplus activation function (AE(Softplus)) and functional autoencoder with Softplus activation function (FAE(Softplus)) using 5 representations for a random test set in Scenario 2.2, when training with 80\% observations (left panel) and 20\% observations (right panel), respectively.}
    \label{fig:sce2.2_Recovery}
\end{figure}

Apart from representation learning, we display the simulated irregularly-spaced functional segments, along with the full curves reconstructed by the nonlinear AE and nonlinear FAE in Figure \ref{fig:sce2.2_Recovery} to reveal the smoothing ability of the FAE. When training with 80\% observations, it is not surprising to observe that the proposed FAE oversteps the classic AE by  generating predominantly smooth curves that effectively capture the entire underlying patterns and primary modes present in the originally observed data. Nevertheless, trajectories obtained through AE exhibit noticeable oscillations and incoherence with numerous accidents protrudes appearing across the entire domain. In the case of training with only 20\% data, as expected, the FAE continues its dominance by showing dramatically leading performances in curve smoothing, highlighting that the FAE with specially designed architecture is able to efficiently learn the representations and simultaneously smooth the unequally-spaced and noisy functional data, even with limited training information.

\section{Real application}
\label{sec:real_application}
\begin{figure}
    \centering
    \includegraphics[width=0.9\textwidth]{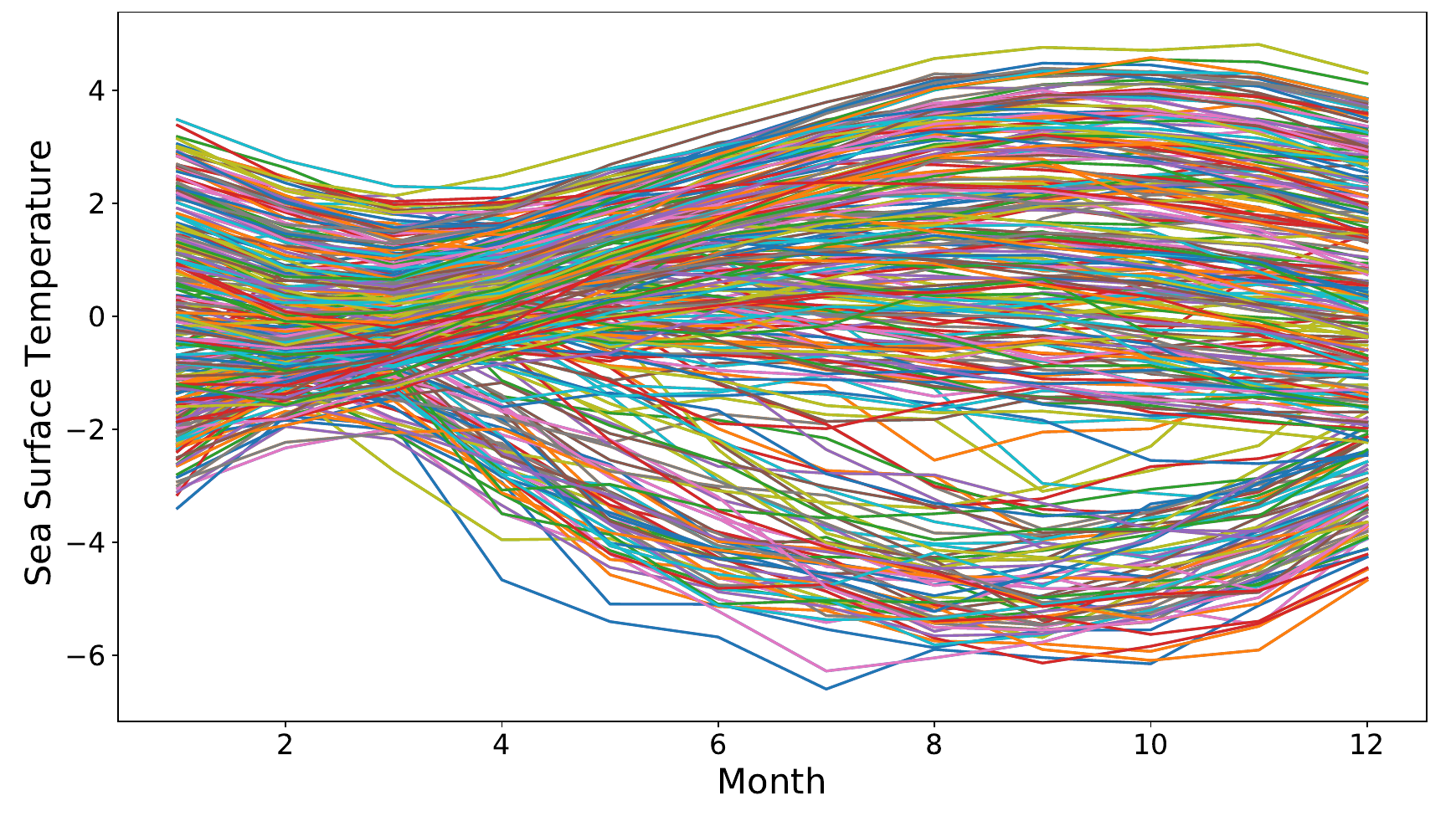}
    \caption{Centered monthly sea surface temperature measured in the ``Ni\~no region'' defined by the coordinates 0-10 degree South and 90-80 degree West.}
    \label{fig:centered_raw_curves}
\end{figure}

To further demonstrate the effectiveness of our method, in this section, we perform the proposed FAE, together  with the conventional FPCA and the classic AE on the El Ni\~no data set which is available in \texttt{R} package \textbf{rainbow} \citep{rainbow}. This data set catalogs the monthly sea surface temperatures originally observed in 4 different locations from January 1950 to December 2006. The temperature curves were discretely measured at 12 evenly spaced time points over the entire interval for every year. We treat the measurements of each calendar year as an independent observation of the true underlying functional curve (varying with time), resulting in a total of 267 observations, and we label the 4 locations with numbers from 1 to 4 randomly. To avoid poor random initialization and obtain stable training process for the NN-based methods, we centre the data by subtracting the sample mean curve across all observations, and a visualization of the centered sea surface temperature curves is provided by Figure \ref{fig:centered_raw_curves}.

We continue to compare the proposed method with two benchmark models, FPCA and classic AE, on their performances in terms of curve reconstruction and representation extraction. We equip the classic AE and the proposed FAE with different combinations of hyper-parameters for a linear mapping path and a potential nonlinear mapping pattern, while FPCA is performed with the focus on measuring the linear relationship. The hyper-parameters for all models in comparison are tuned in advance to yield a fair improvement in their performances during actual training. To reduce the computational cost of tuning process, for each model, we fix the number of representations to be 5 and then perform a grid search over the hyper-parameters of our interests with respect to the loss function by simply building a model for each possible combination of all of the hyper-parameter values provided, and the optimal model architecture combination of hyper-parameters identified by the grid search with 5 representations is further applied to model training with 3 and 8 representations. In the supplementary document, Table S5 provides a summary of the candidate values considered in hyper-parameter tuning for all models, while the details of the optimally identified configurations for models in comparison is narrated in Table S6. We proceed with 20 repetitions of random subsampling validation: randomly dividing the data set into a training set and a test set, with 80\% and 20\% of the total samples assigned to them, respectively. We evaluate the prediction, classification and clustering performance of the proposed methods on the 20 test sets of all the mentioned models using 3, 5 and 8 representations, respectively.

\begin{sidewaystable}
    \centering
    \caption{Means and standard deviations (displayed inside parentheses) of prediction error and classification accuracy of functional autoencode Identity (FAE(Identity)) and Sigmoid activation function (FAE(Sigmoid)), classic autoencoder with Identity (AE(Identity)) and Sigmoid activation function (AE(Sigmoid)) and functional principal component analysis (FPCA) on 20 random test sets with the El Ni\~no data set.}
    \begin{tabular}{@{}c |c | c c c c c @{}}
   \hline
     \multicolumn{2}{c|}{} & \thead{\textbf{FAE (Identity)}} & \thead{\textbf{FAE (Sigmoid)}} & \thead{\textbf{AE (Identity)}} & \thead{\textbf{AE (Sigmoid)}} &  
     \thead{\textbf{FPCA}} \\
     \hline
    \multirow{3}{*}{\thead{MSE$_{\text{p}}$}} & 3 reps  & 0.0616(0.0051) & \textbf{0.0582}(0.0045) & 0.0619(0.0051) & 0.0715(0.0079) &  0.0656(0.0054)\\
    & 5 reps  & \textbf{0.0211}(0.0023) & 0.0226(0.0031)  & 0.0261(0.0052) & 0.0329(0.0042) & 0.0242(0.0031)\\
    & 8 reps  & \textbf{0.0062}(0.0009) & 0.0089(0.0014) & 0.0064(0.0008) & 0.0071(0.0021) & 0.0113(0.0013)\\
    \hline
    \multirow{3}{*}{\thead{P$_{\text{classification}}$}} & 3 reps  & 76.88\%(4.01\%) & \textbf{77.68\%}(5.07\%) & 76.52\%(3.67\%) & 77.14\%(6.05\%) & 77.59\%(4.81\%)\\ 
    & 5 reps & 85.18\%(4.86\%) & \textbf{86.52\%}(4.46\%) & 84.20\%(5.15\%) & 85.71\%(3.48\%) & 84.38\%(5.20\%)\\
    & 8 reps & 85.89\%(4.58\%) & \textbf{87.59\%}(4.67\%) & 85.27\%(3.91\%) & 85.80\%(3.89\%) & 84.81\%(4.50\%) \\
     \hline
    \end{tabular}
    \label{tab:ElNino_result20}
\end{sidewaystable}

Table \ref{tab:ElNino_result20} summaries the performances of all the methods applied for different numbers of representation extracted on curve prediction and classification, using MSE$_{\text{p}}$ and P$_{\text{classification}}$ averaged over 20 random test sets. Apparently, the proposed FAEs consistently and comprehensively outperforms the FPCA and the AE models in terms of both reconstruction and classification, by reaching the lowest prediction error and the highest classification accuracy for all representation attempts. With regard to the predictive performance, the linear FAE retains to be the top performer, closely followed by the nonlinear FAE. On the other hand, the nonlinear FAE continuously oversteps the other models in the competition of classifying curves, exhibiting its advantages in extracting more informative representations. To further confirm this in the context of statistical significance, we conduct two-sided paired t-tests to compare the MSE$_{\text{p}}$ and P$_{\text{classification}}$ of the 20 replicates of the nonlinear FAE to those of the FPCA, and the corresponding p-values are reported in Table S7 in the supplementary document. We observe that the nonlinear FAE remains superior to the FPCA regarding both evaluation metrics, especially when the representation size increases, demonstrating the importance and necessity of the proposed FAE in nonlinear representation learning.

\begin{figure}
    \includegraphics[width=1\textwidth]{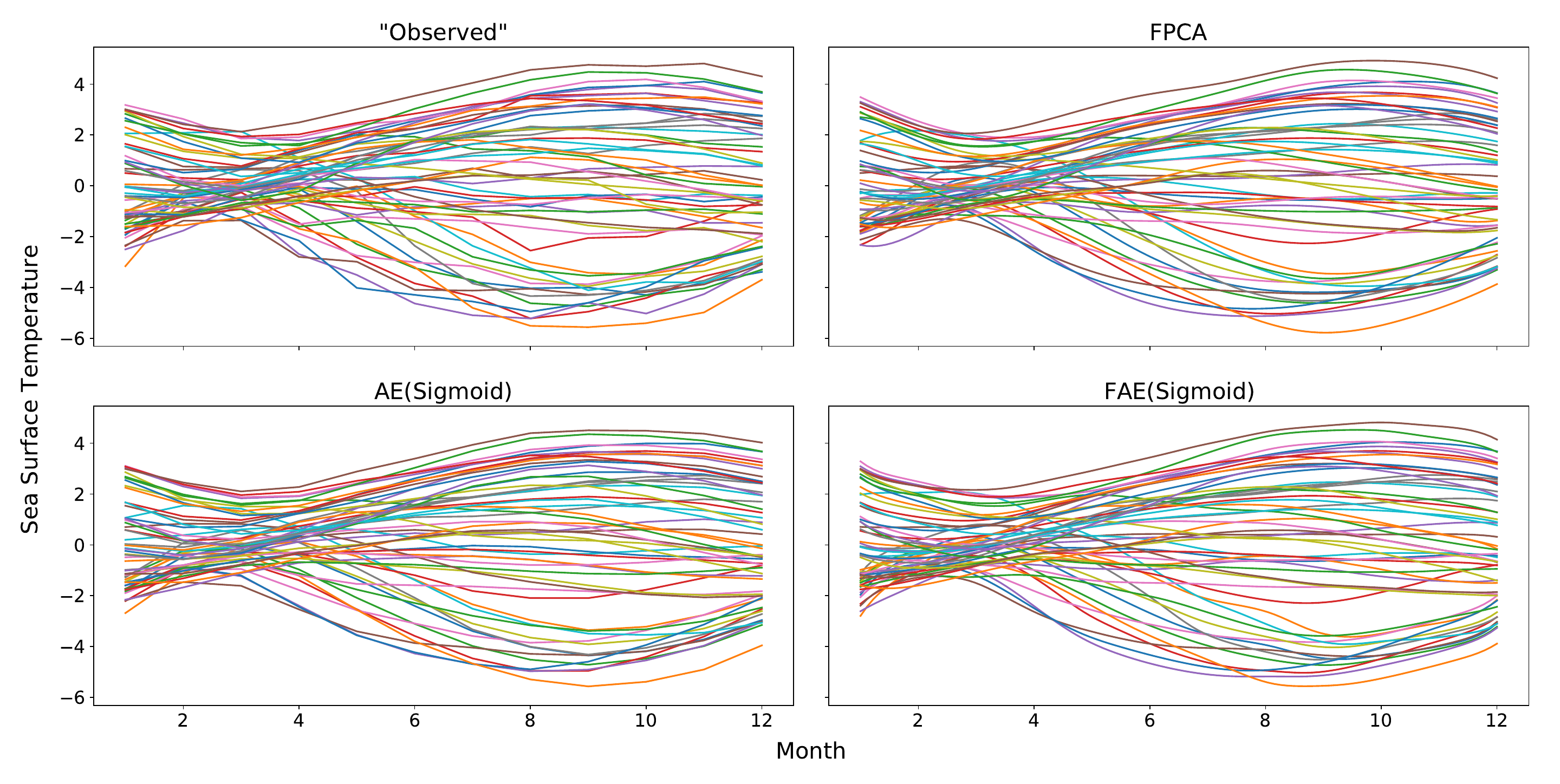}
    \caption{The observed curves and curves recovered by functional principal component analysis (FPCA), classic autoencoder with Sigmoid activation function (AE(Sigmoid)) and functional autoencoder (FAE(Sigmoid)) with 5 representations for a test set of El Ni\~no data set.}
    \label{fig:ElNino_curve_recovery}
\end{figure}

The other highlight of the proposed FAE is its capability of smoothing the originally discrete data. As illustrated in Figure \ref{fig:ElNino_curve_recovery}, the trajectories recovered by using FAE are smooth over the entire domain due to the fact that they are constructed as the linear combination of the neurons in the \textit{coefficient layer} and the basis functions that can be evaluated at any point within the interval of observation. On the contrary, the classic AE can only achieve point-wise prediction at the timestamp with actual observations, resulting in visible discontinuity in the curve reconstruction. Meanwhile, FPCA can also produce smooth prediction but it usually requires the discrete observation to be firstly smoothed.  

\begin{figure}
    \centering
    \includegraphics[width = 1\columnwidth]{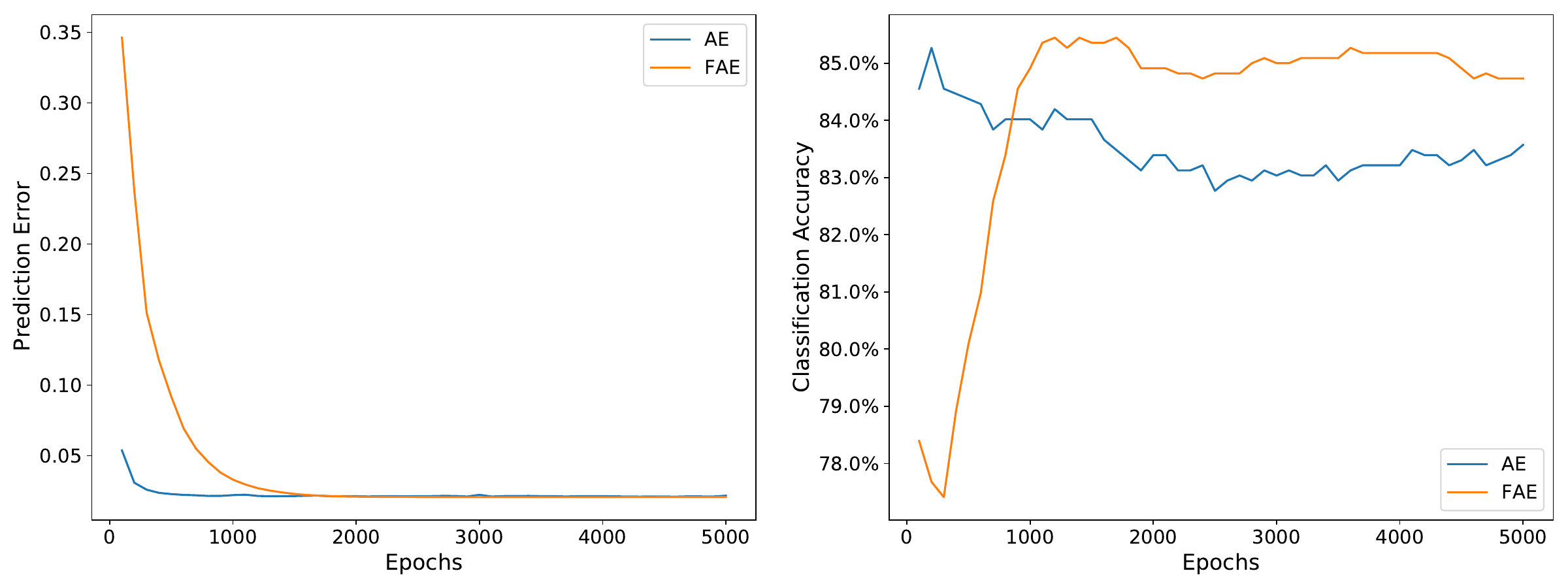}
    \caption{How the averaged prediction error and classification accuracy of functional autoencoder (FAE) and classic autoencoder (AE) with \textbf{Identity} activation function using 5 representations on 20 random test sets of the ElNino data set change with the number of epochs.}
    \label{fig:ElNino_epoch_linear}
\end{figure}

\begin{figure}
    \centering
    \includegraphics[width = 1\columnwidth]{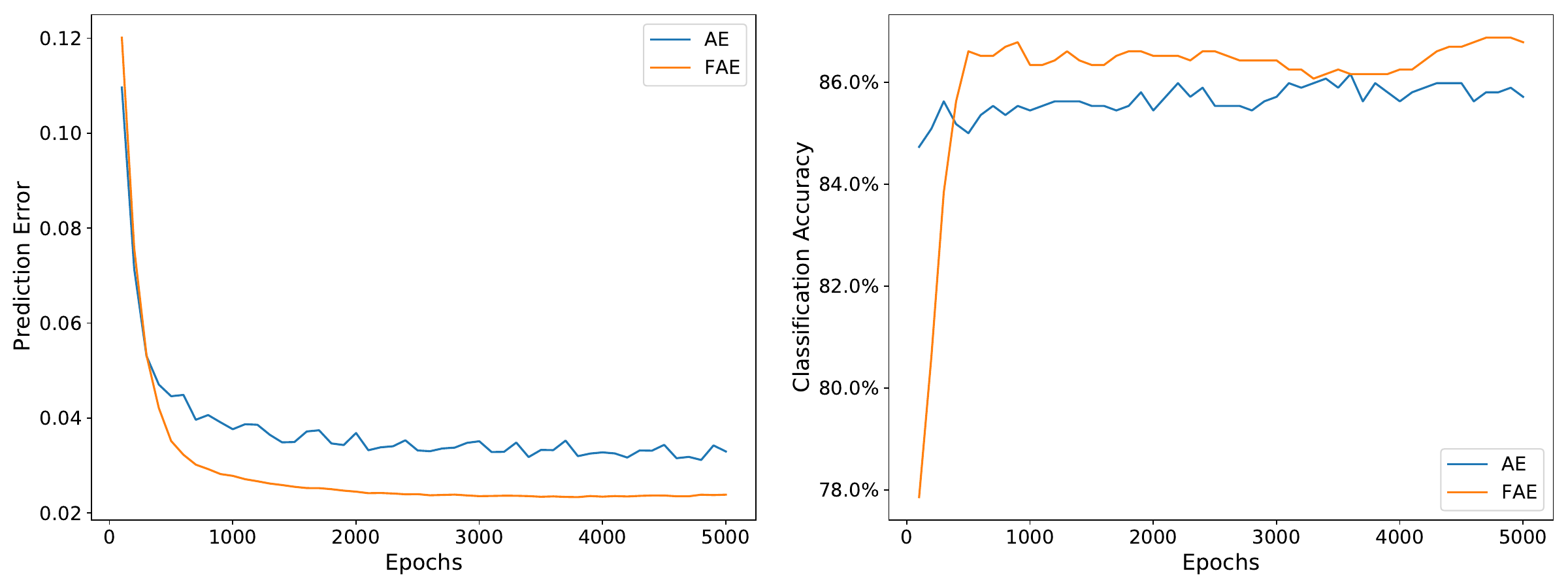}
    \caption{How the averaged prediction error and classification accuracy of functional autoencoder (FAE) and classic autoencoder (AE) with \textbf{Sigmoid} activation function using 5 representations on 20 random test sets of the ElNino data set change with the number of epochs.}
    \label{fig:ElNino_epoch_nonlinear}
\end{figure}

In addition, FAE surpasses AE by fast converging to a similarly low prediction error but with a resulting higher classification accuracy in both linear and nonlinear scenarios, as displayed in Figure \ref{fig:ElNino_epoch_linear} and Figure \ref{fig:ElNino_epoch_nonlinear}, demonstrating its high efficiency in extracting meaningful features and potential advantage in saving computational cost. It is noteworthy that the model configuration for the nonlinear AE is simpler than that of the nonlinear FAE, which brings benefits to the speed of nonlinear AE in training loss convergence. 

The given results might imply that, for the El Ni\~no data, the true relationship between the functional space to representation space for the sea temperature curves can be more accurately learned and revealed through a nonlinear mapping path, with the resulting nonlinear representations carrying more valuable information ready for further statistical analysis.

\section{Concludsion}
\label{sec:conclusion}
In this work, we introduced autoencoders with a new architecture design for discrete functional observations targeting at unsupervised representation learning and direct curve smoothing concurrently. The deterministic \textit{feature layer} added to the encoder reduces the computational effort and enhances the model robustness in learning performances, while the additional \textit{coefficient layer} similarly incorporated into the decoder allows the usage of back-propagation in model training and allows the decompression from scalar neurons to functional curve in a linear manner. Additionally, we implemented our proposed FAE in a way that accommodates both regularly and irregularly functional data with flexible necessity on the size of training data for achieving satisfactory performances. Through simulation studies and real applications, we demonstrated that the method we proposed is superior to other linear representation method for functional data, i.e., FPCA, under nonlinear scenarios and retains competitive performances in linear cases. Moreover, the numerical experiments endorse that our model can be a dramatic improvement over the classic AE in terms of computational efficiency by generalizing and converging rapidly even with reduced training observations and limited computing time.

Nevertheless, the developed method depends on numerous hyper-parameters, including number of hidden layers, number of neurons in each hidden layer, training optimizer, etc., and unfortunately conducting a grid search on that space can be time-consuming. It is necessary to bring up that the performance of the FAE varies from one hyper-parameters configuration to another, and the randomness in initializing network parameters will introduce more variance to the results across training replicates. In contrast, the FPCA can be effortlessly fitted with only one hyper-parameter necessitating predetermination, but in sacrifice of the ability to accurately capturing the learning maps in nonlinear situations. Another weakness of our approach is its inability to process multidimensional functional data in its current form. Therefore, in a future work we could extend the current network architecture to a more dynamic architecture allowing discrete multivariate functional data. This might be achieved by introducing micro-neural networks \citep{Micro_NN, AdaFNN} to replace the neurons in the \textit{feature layer} and the \textit{coefficient layer}. Furthermore, an analogous architecture of our proposed FAE can be implemented to tackle nonlinear functional regression problems with both a functional predictor and a functional response.

\bibliography{sn-bibliography}

\end{document}